%% file: main.tex
\documentclass{article}
\usepackage[T1]{fontenc}

\usepackage{microtype}
\usepackage{graphicx}
\usepackage{subcaption}
\usepackage{booktabs} 

\usepackage{hyperref}

\usepackage[preprint]{icml2026}

\usepackage{amsmath}
\usepackage{amssymb}
\usepackage{mathtools}
\usepackage{amsthm}

\usepackage[capitalize,noabbrev]{cleveref}

\theoremstyle{plain}

\theoremstyle{definition}

\theoremstyle{remark}

\usepackage[textsize=tiny]{todonotes}

\icmltitlerunning{Decoupling Generalizability and Membership Privacy Risks in Neural Networks}

\begin{document}

\twocolumn[
  \icmltitle{Decoupling Generalizability and Membership Privacy Risks in Neural Networks}



  \icmlsetsymbol{equal}{*}

  \begin{icmlauthorlist}
    \icmlauthor{Xingli Fang}{ncsu}
    \icmlauthor{Jung-Eun Kim}{ncsu}
  \end{icmlauthorlist}

  \icmlaffiliation{ncsu}{Department of Computer Science, North Carolina State University}

  \icmlcorrespondingauthor{Jung-Eun Kim}{jung-eun.kim@ncsu.edu}

  \icmlkeywords{Machine Learning, ICML}

  \vskip 0.3in
]



\printAffiliationsAndNotice{}  

\begin{abstract}
A deep learning model usually has to sacrifice some utilities when it acquires some other abilities or characteristics. 
Privacy preservation has such trade-off relationships with utilities. 
The loss disparity between various defense approaches implies the potential to decouple generalizability and privacy risks to maximize privacy gain.
In this paper, we identify that the model's generalization and privacy risks exist in different regions in deep neural network architectures. Based on the observations that we investigate, we propose \textit{Privacy-Preserving Training Principle} (PPTP) to protect model components from privacy risks while minimizing the loss in generalizability. Through extensive evaluations, our approach shows significantly better maintenance in model generalizability while enhancing privacy preservation.
\end{abstract}


\input{sec/intro}

\input{sec/relatedwork}
\input{sec/where}
\input{sec/approach}

\input{sec/exp}
\input{sec/con}
\bibliography{main}
\bibliographystyle{icml2026}

%
%

\end{document}

%% file: sec/intro.tex
\section{Introduction}


A machine learning model acquires accurate recognition abilities by learning how to fit the training data points. However, even if a model shows a good performance for an objective of a given task, it may suffer from the risk of data privacy leakage, especially in privacy-sensitive applications or systems. The concern about this issue is also raised by several existing studies: \cite{devansh2017meminnn, chatterjee2018learnmemorization} showed neural networks' potential privacy risks via fitting on random data. \cite{shokri2017membership} showed that privacy risks are widely posed in neural networks by showing the possibility of black-box membership inference attacks (MIAs). 
Further studies \cite{choquette2021labelonlymia,del2022leveraging} showed that a model has significant behavioral differences between the training and testing data in various aspects, especially in robustness. Also,
\cite{stephenson2021geometry} found that the model's memory of training data points becomes solid as training progresses. Besides, \cite{yuan2022samia} observed that pruning does not help neural networks mitigate these behavioral discrepancies, identifying that privacy risks are not evenly distributed across the model. All of these prior observations and knowledge indicate that machine learning models have a strong memory for data, which enables the model to achieve near-perfect performance on the training dataset.

Since a well-trained model will retain a lot of data traces, deploying a model in a privacy-sensitive system requires particular caution. This risk occurs not only in classification models \cite{shokri2017membership, song2021systematic, choquette2021labelonlymia} but also in some other machine learning domains, such as generative models \cite{chen2020ganleaks} or transfer learning \cite{zou2020miatransfer, wu2024rethinkmiatransfer}, etc. Besides, models could leak privacy in various ways, e.g., membership inference attacks \cite{shokri2017membership}, model inversion attacks \cite{fredrikson2015mi}, and model extraction attacks \cite{tramer2016me}. This universality makes the output of any model pose the potential risk of privacy leakage. Therefore, understanding the sources of privacy risks and how to rectify them is the key to strengthening the model to be trustworthy.

In this paper, we study where and how layer-level privacy leakage occurs in neural network architectures. We empirically identify that the model's generalizability and privacy risk are separable. Then, we propose Privacy-Preserving Training Principle (PPTP) to minimize negative impacts on the model utility during training with privacy defense approaches. 
Here is a brief overview of our novel observations and contributions:
\noindent
\begin{itemize}
    \item We structurally and precisely investigate where and how the machine learning model produces privacy risks, and identify that \textbf{\emph{privacy risks and generalizability occur in different regions in a model.}}
    \item We propose a new cost-effective training paradigm for utility-privacy trade-offs. It \textbf{\emph{decouples utility and privacy}} into two separate parts, enabling the model to outperform existing privacy defense approaches.
    \item Our approach bridges the gap between generalizability and privacy training, empowering the model to choose its proper training approaches (for both utility and privacy) based on its application contexts and helping the model overcome the limitations of the existing privacy defense approaches.
\end{itemize}

%% file: sec/relatedwork.tex
\section{Related Work}

There have been various studies \cite{fang2025trustworthyaisafetybias} to prevent the ML model from privacy risks. DP-SGD \cite{abadi2016dpsgd} tried to train the model with a less fitting degree via additional noises in the optimizer. 
\cite{nasr2018advreg} developed an adversarial mechanism to help the model obtain aligned predictions. 
However, both require significantly increasing training costs due to their limitations in parallel computations.
\cite{jia2019memguard, yang2023purifier} proposed decorators to reproduce better-aligned predictions without retraining the model.
\cite{shejwalkar2021dmp} tried to mitigate the privacy risk via knowledge distillation. \cite{tang2022selena} further improved the utilization rate of training data via ensemble-based knowledge distillation.
\cite{li2021mixupmmd} attempted the alignment between training and non-training accuracy during training.
\cite{chen2022relaxloss} proposed an efficient training paradigm with effective privacy-preserving ability.
\cite{fang2024crl, fang2024srcm} discussed the privacy issues in the bottleneck layer. 
\cite{liu2024ccl} observed the convexity of loss functions is a factor of privacy leakage and tried to mitigate it with a concave term.
Although the conclusions are not entirely consistent, \cite{wang2021pruning, yuan2022samia} explored the impact of network pruning on privacy risks.
\cite{kaya2021whendataaug, yu2021howdoesdataaug} explored which data augmentation techniques are beneficial to privacy.
\cite{tan2023blessing} pointed out that higher model dimensions are possibly more privacy-risky.
\cite{li2024mist} tried to focus on the most privacy-risky data points.
\cite{carlini2022privacyonion} found that a model always experiences varying degrees for different data points in traditional training settings.
\cite{chen2023hamp} tried to mitigate the privacy risks by reducing overconfidence during both training and inference.
\cite{zhang2024archprivacy} found some components lead to machine learning models at severe privacy risks.
\cite{zhao2025does_syn_data_protect_privacy} evaluated and observed the privacy flaws of the prior data synthesis approaches.
\cite{shang2025defending_mias_iteratively_prune_dnn} made use of the memorization destroying during iterative pruning to retrain the model on the identified privacy-risky samples in a less overfitting way.
\cite{fang2026learnability} evaluated the privacy vulnerabilities among weights.

Despite considerable progress, it still remains a work in progress. As with other characteristics (e.g., Robustness \cite{goodfellow2015advattack, Mulchandani2025ICLR}, Fairness \cite{mehrabi2021fairnesssurvey,Cuong2022NeurIPS, Bellam2025QFITMLNeurIPS, Bellam2025QFairness}, and Data Unlearning \cite{bourtoule2021mul}), models inevitably have to pay the price of utility for privacy with current approaches. Sometimes, this expense is even too high for the model to maintain reasonable performance due to the inherent characteristics of the dataset \cite{carlini2022lira}. We question the preconceived view kept by current privacy-preserving approaches that the model is a privacy-risky entity and explore where the model produces the privacy risks in the next section.

%% file: sec/where.tex
\begin{figure*}[t]
     \centering
     \includegraphics[width=1.\linewidth]{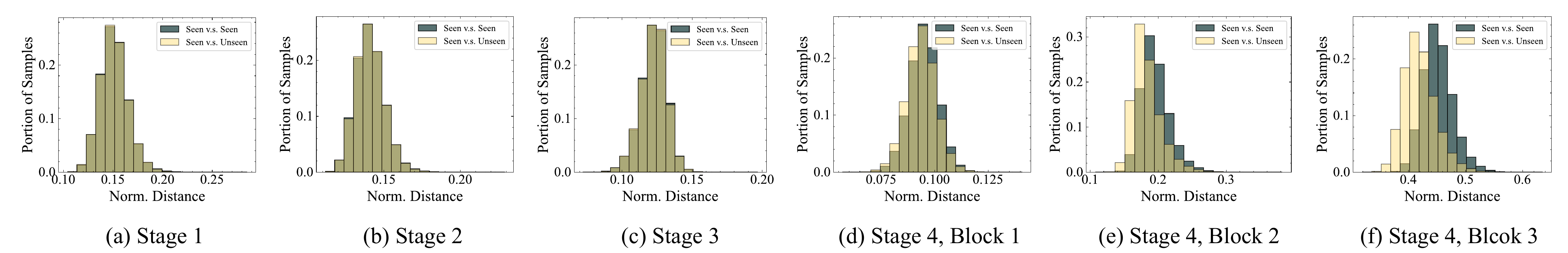}
\caption{The sample-level feature map differences (norm. distance on x-axis). No disparity is observed in Stage 1--3, whereas gradually increasing disparity is observed within Stage 4 (ResNet152, TinyImageNet, data augmented)}
\label{fig:resnet152_seen_vs_unseen}
\end{figure*}

\section{Does Prediction Disparity Exist Everywhere?}
\subsection{Why the Question Matters}
Recent non-decorator defense approaches \cite{abadi2016dpsgd, nasr2018advreg, shejwalkar2021dmp, chen2022relaxloss, tang2022selena} usually train a model from scratch while decorator approaches \cite{jia2019memguard, yang2023purifier} usually add extra filters to the model externally rather than training the model itself. A direct advantage of training from scratch is that it is implementation-friendly. However, training the whole model could lead to an unnecessary utility loss as we see a discrepancy between adversarial training on the whole model \cite{shafahi2019advfree} and training the last layer only \cite{kirichenko2023robustlast}. Then, it gives us a question: \textbf{Does prediction disparity exist everywhere}? In this section, we discuss the correlation between privacy-risk and other attributes of machine learning models, such as feature map size, channel size, and depth.

It is known that the model's generalizability is affected by the depth \cite{baldock2021lendifficulty}. That is, a model gradually obtains generalizability layer by layer while learning samples with various difficulty levels. If privacy risk accompanies the learning of different difficulty-level samples, the privacy risk should exist in each layer and ``\emph{gradually}'' show more and more prediction disparity. Meanwhile, \cite{kirichenko2023robustlast} showed that spurious correlations can be mitigated by retraining only the last layer, hinting that different layers may not exhibit the same characteristics with regard to generalizability. Such recent insights challenge us to locate where a disparity between generalizability and privacy risks starts occurring.

\subsection{Sample-Level Measurement Design}
Unlike the logits produced by the classification layer, there is no direct way to know what a confident feature map should look like. Therefore, we develop a method to examine distribution disparity. Let $D_{all}$ denote the entire training set. We split $D_{all}$ into two halves: $D_{h1}$ and $D_{h2}$. Then, we train two models, $M_{all}$ and $M_{h1}$, with the same architecture and configurations with training sets $D_{all}$ and $D_{h1}$, respectively. Therefore, every data point $x_1 \in D_{h1}$ is seen data for both two models, while $x_2 \in D_{h2}$ is seen data for only $M_{all}$ but unseen data for $M_{h1}$. 

Then, we compute the feature map differences using Euclidean distance $\texttt{Dist}(p, q) = \sqrt{\sum_{i=1}^{d}(p_i-q_i)^2/d}$, where $d$ is the number of dimensions of the feature maps and $1/\sqrt{d}$ is an extra normalized item, while $p$ and $q$ denote the feature maps produced by $M_{all}$ and $M_{h1}$ on the same input, respectively. 
If a layer does not produce prediction disparity, then the feature map distance distributions on $D_{h1}$ and $D_{h2}$ should be equivalent, and vice versa.

\subsection{Empirical Verification}
\begin{figure}[t]
     \centering
     \includegraphics[width=1.0\linewidth]{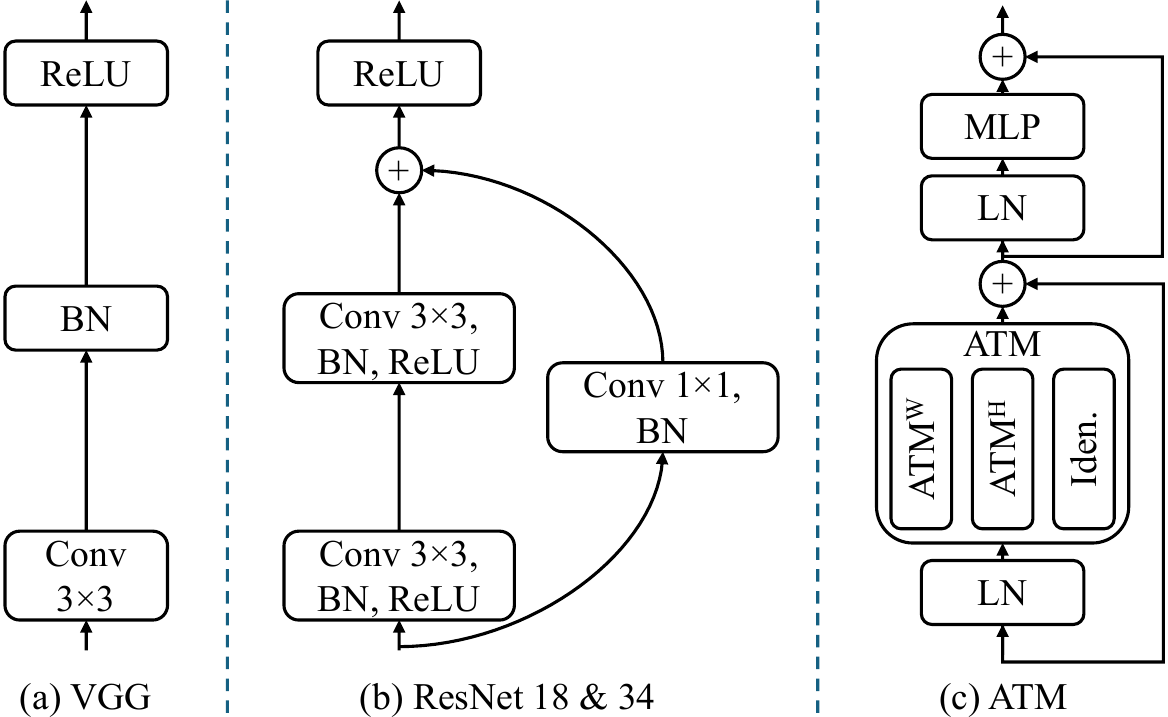}
\caption{Overview of the three architectures' backbone modules.}
\label{fig:draw_architecture}
\vskip -0.15in
\end{figure}
First of all, we need to check if the disparity happens in all layers or not. Fig.~\ref{fig:resnet152_seen_vs_unseen} plots sample-level feature map differences of seen and unseen data (norm. distance on x-axis) where we find that the disparity \textbf{\emph{abruptly}} starts in stage 4. In other words, there is only a small portion (the 4th stage contains only 3 blocks while the entire model contains 33 blocks) of ResNet152 that leaks privacy, identifying that the model is not globally privacy-risky, but \emph{regionally}. However, what conditions cause disparity is not clear yet. Hence, we explore different factors' impacts on disparity in the rest of this section. We explore these factors using three different architectures as shown in Fig.~\ref{fig:draw_architecture}: VGG \cite{simonyan2015vgg}, ResNet \cite{he2016resnet}, and Active Token Mixer (ATM) (which is a transformer-type architecture) \cite{wei2023atm}. In the three architectures, ResNet and ATM have residual connections, while VGG does not. ATM has self-attention-type computation modules, which differentiates it from VGG and ResNet.

\begin{figure}[t]
     \centering
     \includegraphics[width=1.\linewidth]{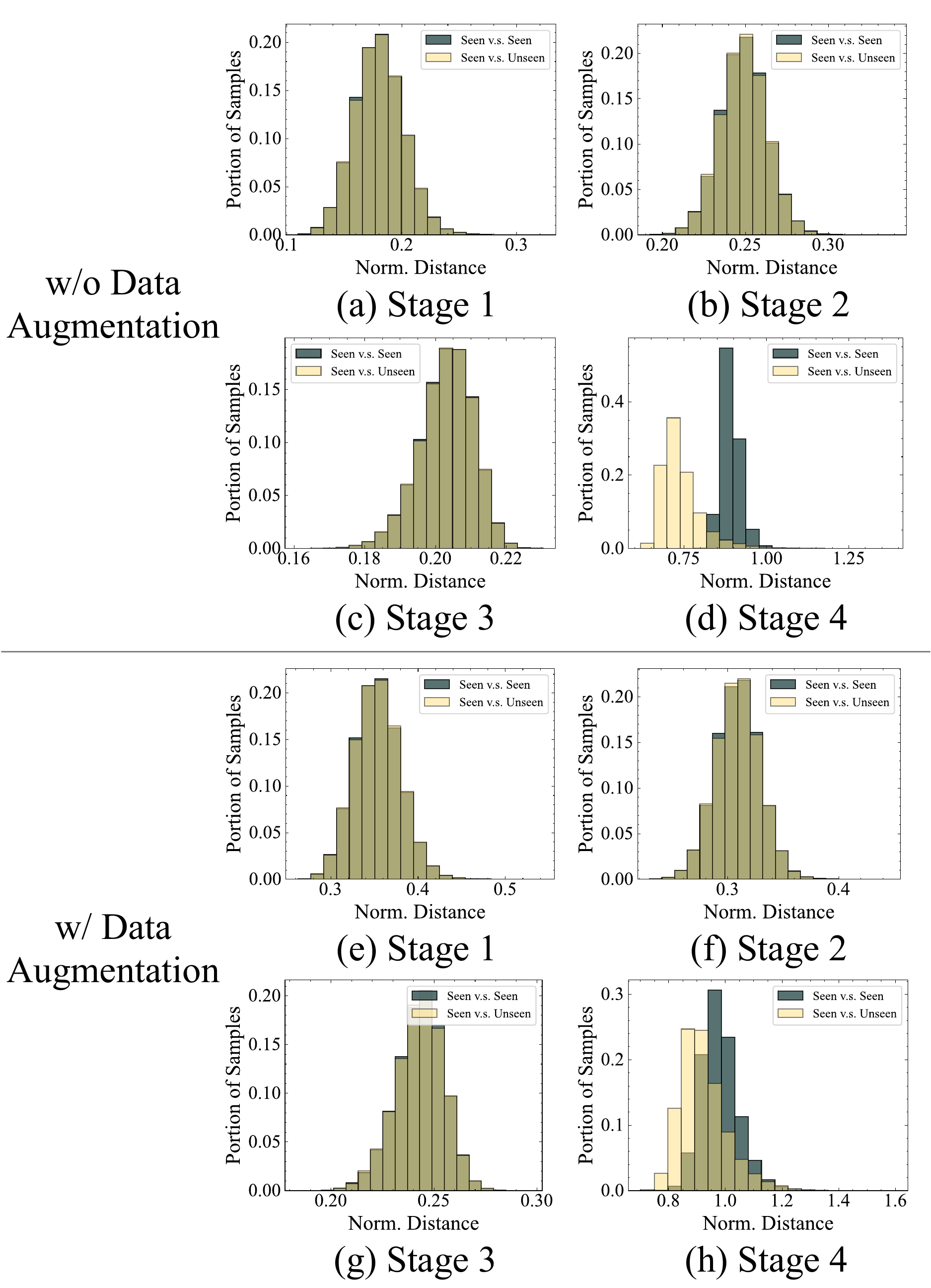}
\caption{Comparison of models trained with and without data augmentation. (ResNet18, TinyImageNet)}
\label{fig:aug_vs_noaug}
\vskip -0.15in
\end{figure}

\paragraph{Data Augmentation} 
\cite{kaya2021whendataaug, yu2021howdoesdataaug} found some data augmentation techniques are beneficial to the model's privacy. Figure~\ref{fig:aug_vs_noaug} shows that data augmentation does help mitigate disparity when it occurs (please compare (d) and (h)). However, whether with or without data augmentation, disparity happens in stage 4, which exhibits that data augmentation has no impact on ``where'' disparity happens.

\begin{figure*}[t]
    \centering
    \includegraphics[width=.8\linewidth]{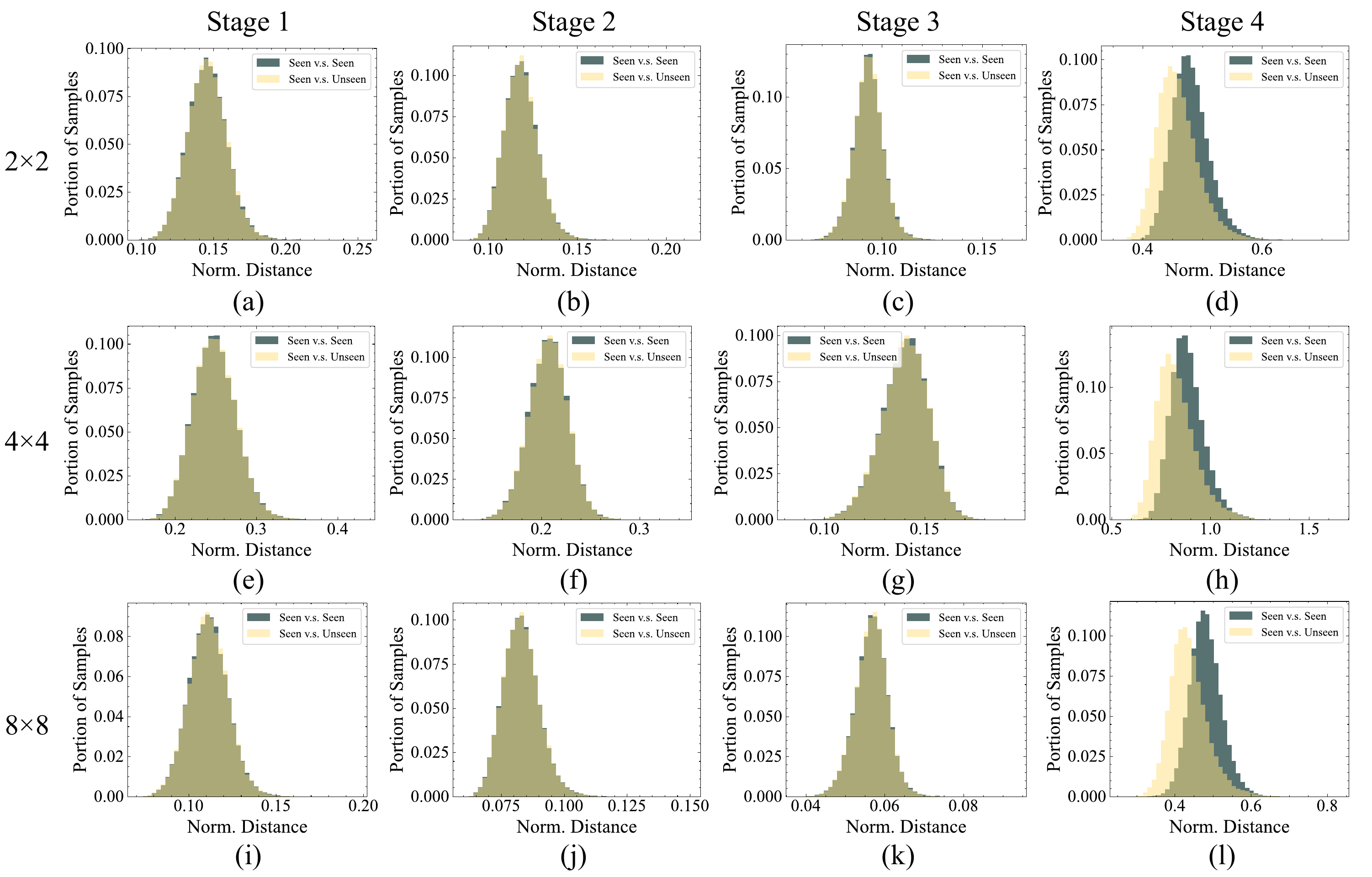}
    \caption{Comparison of ResNet18 with different feature map sizes in the 4th stage. (TinyImageNet, data augmented)}
    \label{fig:resnet_feasize}
\end{figure*}
\begin{figure}[t]
    \centering
    \includegraphics[width=1.\linewidth]{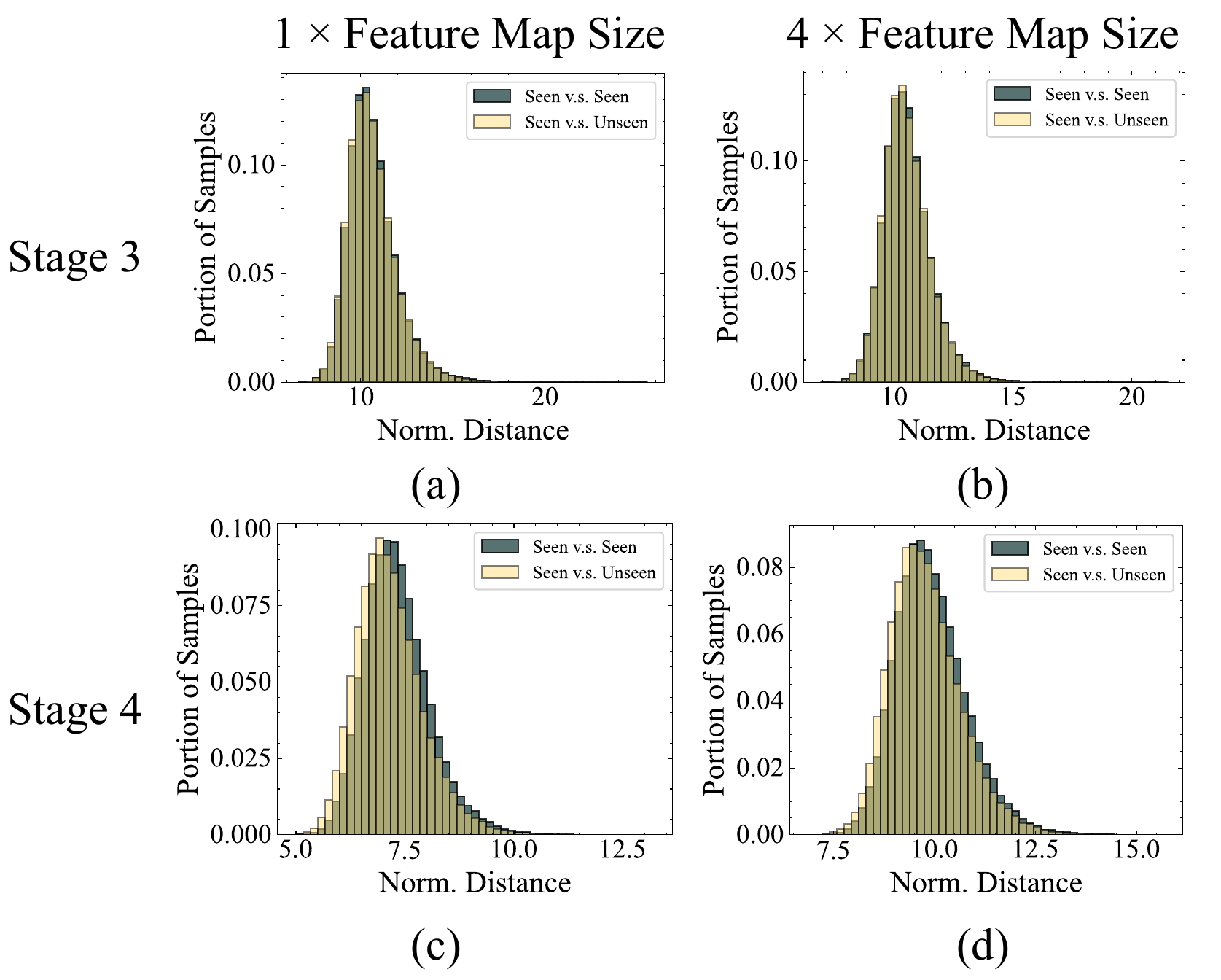}
    \caption{Comparison of ATM-XT in various channel sizes at the 3rd \& 4th stage. (TinyImageNet, data augmented).}
    \label{fig:atm_feasize}
\end{figure}
\begin{figure*}[t!]
     \centering
     \includegraphics[width=1.\linewidth]{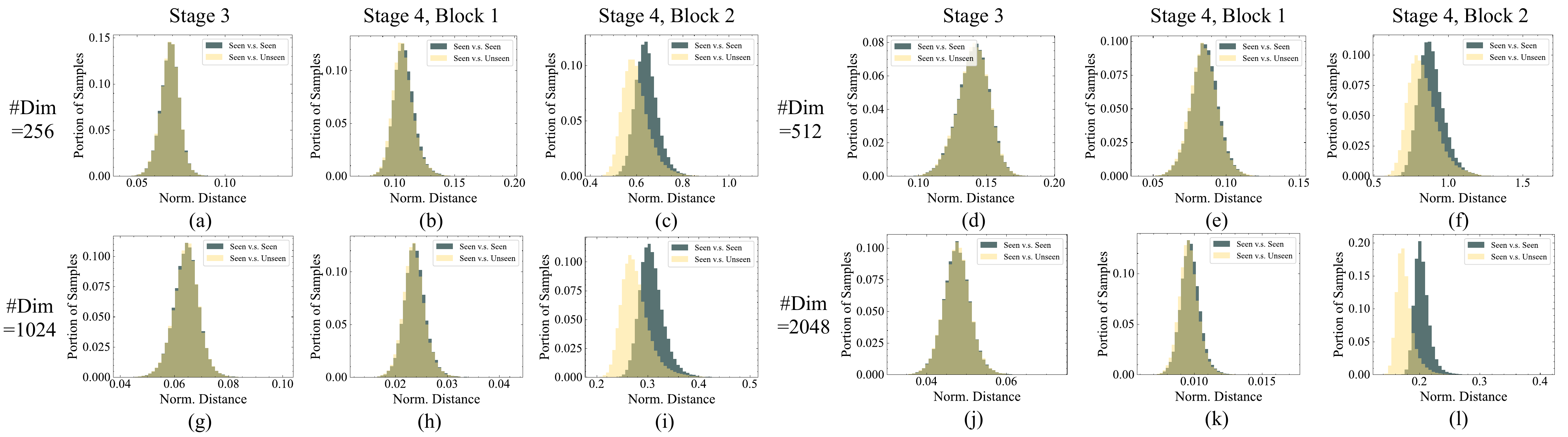}
\caption{Comparison of ResNet18 in various channel sizes at the 4th stage. (TinyImageNet, data augmented)}
\label{fig:resnet_csize}
\end{figure*}

\paragraph{Feature Map Size}
One significant change along with stages is the feature map size. After each down-sampling layer at the head of each stage in the ResNet (similar to many widely used architectures such as CNN, ViT, Mixer, etc.), the feature map's width and height are halved. To study how generalizability and disparity change along with feature map size, the feature map size at the 4th stage of ResNet18 is enlarged and reduced (originally $4\times 4$). By considering both Figure~\ref{fig:resnet_feasize} and Table~\ref{tab:resnet_feasize} together, we find that, while enlarging the feature map also has no impact on where the disparity happens, it enhances the model's generalizability while not significantly exacerbating the disparity. The same trends are also observed in ATMs (refer to Figure~\ref{fig:atm_feasize} and Table~\ref{tab:atm_feasize}.
As shown in Fig.~\ref{fig:atm_feasize}, enlarging the feature map size does not affect when the privacy risks occur in ATM-XT. Their performance on test set can be seen in Table~\ref{tab:atm_feasize}.

\begin{figure}[t]
     \centering
     \includegraphics[width=1.0\linewidth]{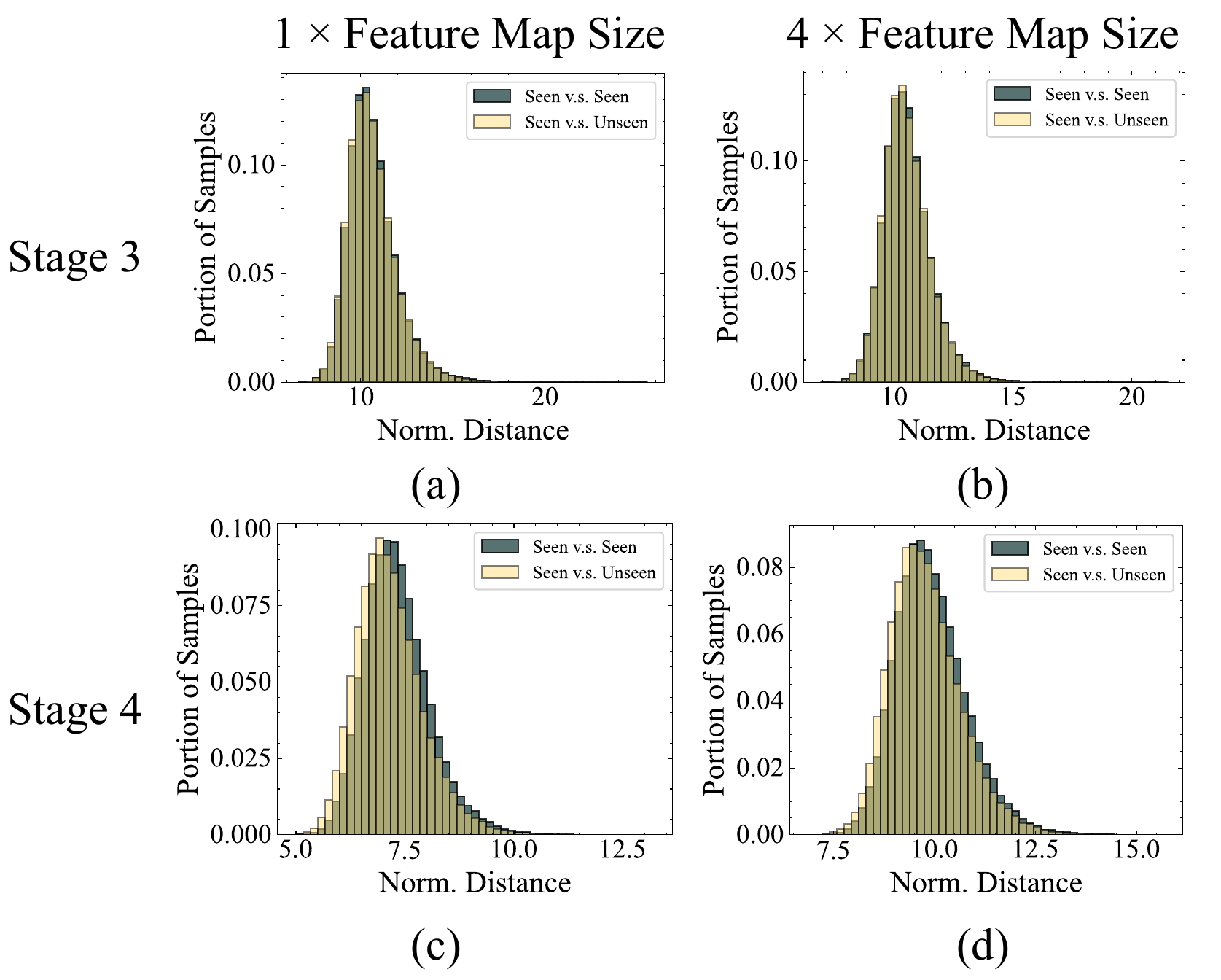}
\caption{Comparison of ATM-XT in various channel sizes at the 3rd \& 4th stage. (TinyImageNet, data augmented).}
\label{fig:atm_feasize}
\end{figure}

\begin{table}[t]
\centering
\caption{The performance comparison among different feature map sizes at the 4th stage. (ResNet18, TinyImageNet, data augmented)}
\vskip 0.1in
\resizebox{.75\linewidth}{!}{
\begin{tabular}{c|cccc}
  \toprule
              & \multicolumn{3}{c}{Feature map size} \\
               \cmidrule(r){2-4}
   Accuracy (\%) & $2\times2$ & $4\times4$ & $8\times8$ \\
  \midrule 
      Train & 99.98 & 99.98 & 99.98  \\
      Test  & 52.32 & 54.58 & 61.53  \\
  \bottomrule
\end{tabular}}
\label{tab:resnet_feasize}
\end{table}

\begin{table}[t]
\centering
\caption{The performance comparison among different feature map sizes at the 4th stage. (ATM-XT, TinyImageNet, Data Augmentation)}
\vskip 0.1in
\resizebox{1.0\linewidth}{!}{
  \begin{tabular}{c|cccc}
    \toprule
     Accuracy (\%) & $1\times$ Feature Map Size & $4\times$ Feature Map Size  \\
    \midrule 
        Train & 99.97 & 99.98  \\
        Test  & 36.28 & 47.29 \\
    \bottomrule
  \end{tabular}}
\label{tab:atm_feasize}
\end{table}

\begin{table}[t]
\centering
\caption{The testing accuracy (\%) comparison among different training data and different channel sizes at the 4th stage. (ResNet18, TinyImageNet, data augmented)}
\vskip 0.1in
  \centering
  \resizebox{.85\linewidth}{!}{
  \begin{tabular}{c|cccc}
    \toprule
                    & \multicolumn{4}{c}{Channel size} \\
                    \cmidrule(r){2-5}
    Accuracy (\%)       & 256   & 512   & 1024  & 2048 \\
    \midrule 
        Test & 52.52 & 54.58 & 53.99 & 55.03     \\
    \bottomrule
  \end{tabular}}
  \label{tab:resnet_csize}
\end{table}

\paragraph{Channel Size}
One common design trend in deep learning in the recent decade is increasing channel size. As shown in Fig.~\ref{fig:resnet_csize} and Table~\ref{tab:resnet_csize}, the disparity becomes more and more significant as channel size increases while changes in testing accuracy are not as significant as the changes by enlarged feature map size (refer to Table~\ref{tab:resnet_feasize}) at 4th stage. That is, too many channel sizes make the model prone to produce disparity, leading to more severe privacy risks. Therefore, it is not worth designing too many channels from the point of view of utility-privacy trade-offs.



\begin{figure*}[t!]
     \centering
     \includegraphics[width=1.\linewidth]{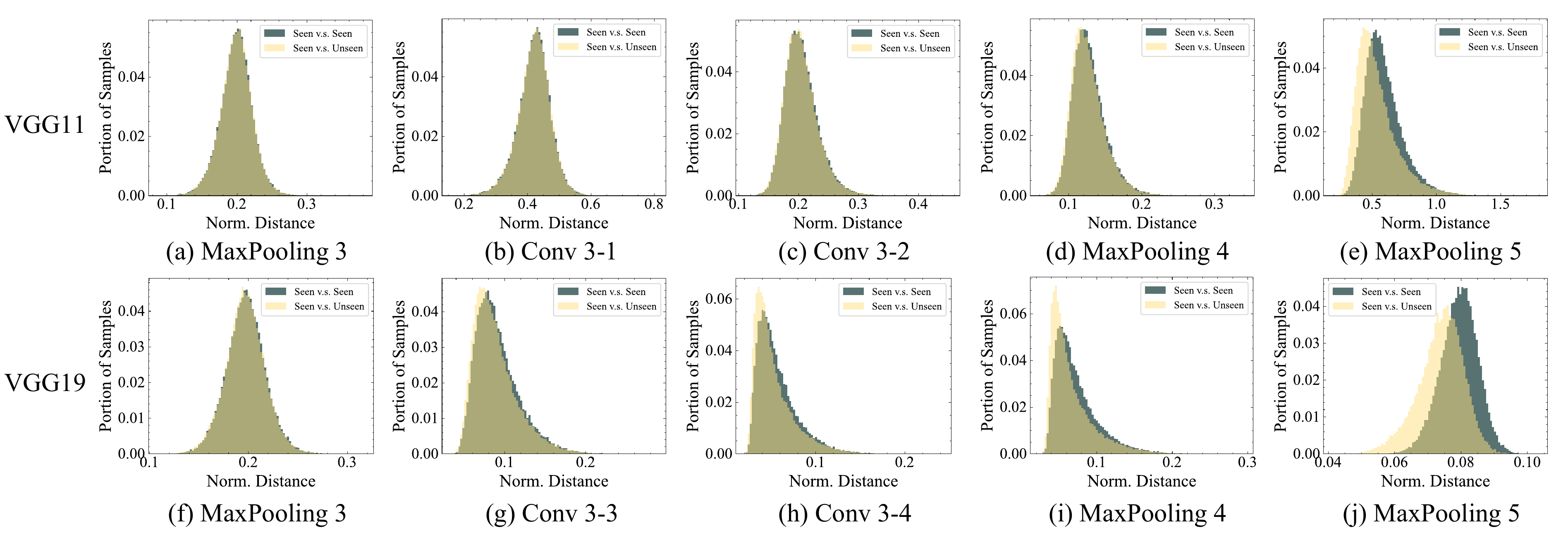}
\caption{The sample-level feature map differences measurement. The disparity is observed earlier in VGG19. (VGG11\&19, TinyImageNet, data augmented)}
\label{fig:vgg_seen_vs_unseen}
\end{figure*}

\paragraph{Depth}
\begin{figure*}
\centering
\subfloat[ATM]{
\label{fig:atm_depth_acc}
   \includegraphics[width=.26\linewidth]{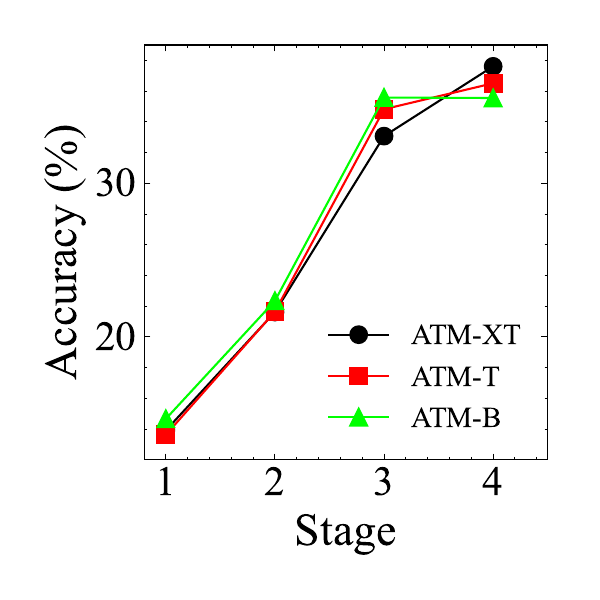}
}
\subfloat[VGG]{
\label{fig:vgg_depth_acc}
   \includegraphics[width=.26\linewidth]{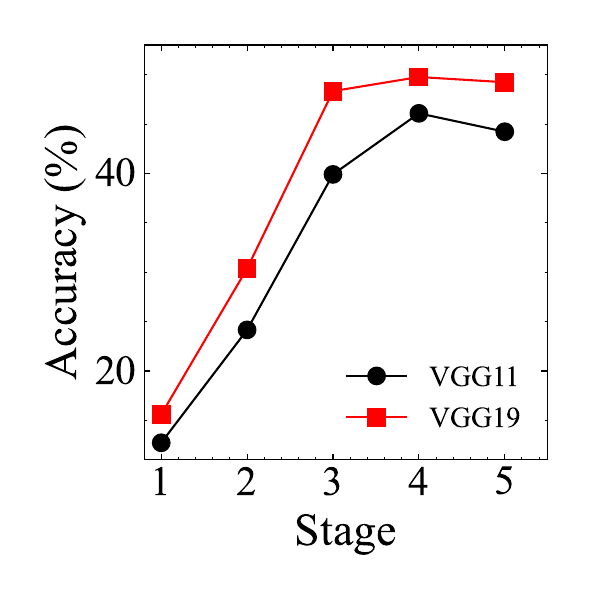}
}
\subfloat[ResNet]{
\label{fig:resnet_depth_acc}
   \includegraphics[width=.26\linewidth]{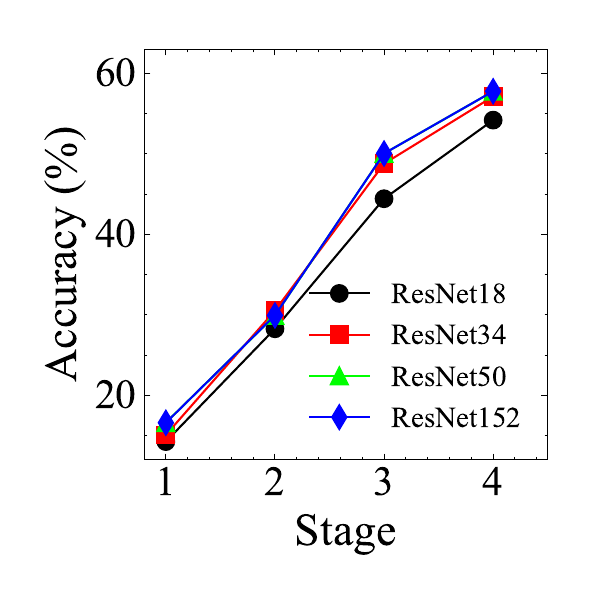}
}
\caption{Testing accuracy changes along with stages using SVM. (TinyImageNet)}
\label{fig:acc_depth}
\end{figure*}
As observed above, since feature map size and channel size do not impact where privacy risk occurs, now we investigate the depth of the model. First, we use a support vector machine (SVM) \cite{cortes1995svm} to classify the features extracted in each stage (see Figure~\ref{fig:acc_depth}). We notice that some stages have limited (or even no) benefits to the model's generalizability. The testing accuracy in all models keeps consistently increasing in the early stages (stages 1, 2, and 3 in Figure~\ref{fig:acc_depth}). However, this growth rate starts to level off after stage 3 to various degrees - some of them slow down (ResNet), and others stagnate or even deteriorate (ATM and VGG). Please note that this stagnation of growth perfectly matches where the privacy risk occurs (e.g., Figure~\ref{fig:vgg_seen_vs_unseen} \& Figure~\ref{fig:acc_depth}(b)). It explains that when the model cannot learn sufficient features to be generalized enough as much as its capacity, it instead learns many ineffective features. These features cannot help generalizability but can help the model fit better on the train set. Therefore, these features must be privacy-risky since they are only valid on the train set.

\begin{figure*}[t]
     \centering
     \includegraphics[width=1.\linewidth]{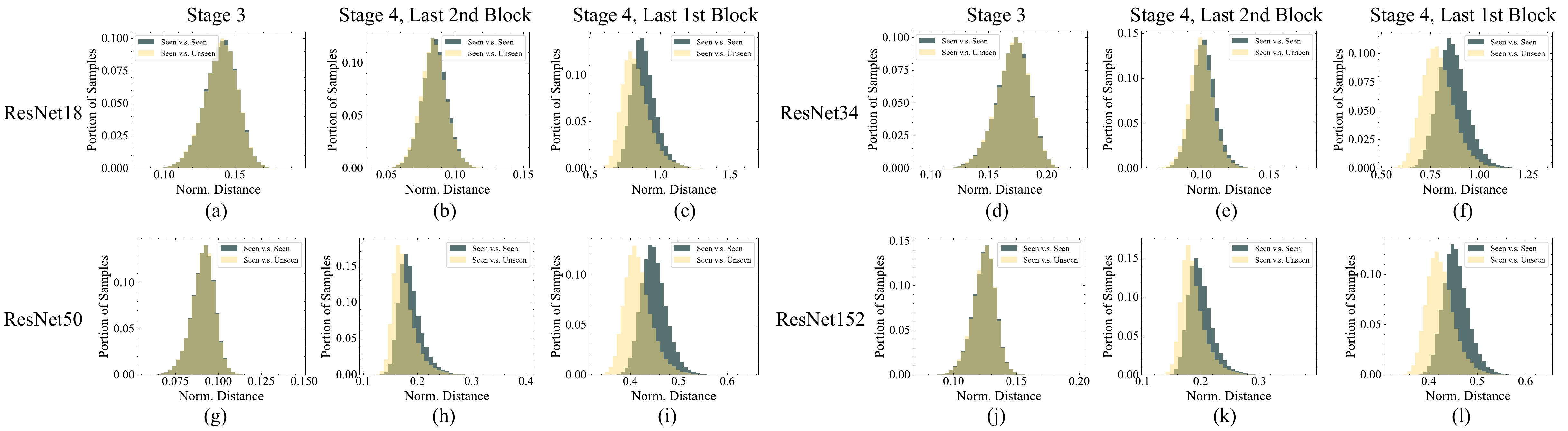}
\caption{Comparison of feature map differences among different ResNet depths.}
\label{fig:resnet_depth}
\end{figure*}

Besides, another notable point is that VGG shows different trends (privacy risks occur at different layers in VGG11\&19, see Figure~\ref{fig:vgg_seen_vs_unseen}) from ATM's case (privacy risks always start at the 3rd stage, see Figure~\ref{fig:atm_depth}) and ResNet's case (privacy risks always start at the 4th stage, see Figure~\ref{fig:resnet_depth}). That is because VGG11 can maintain the growing speed of generalizability between stages 3 and 4 while VGG19 cannot. In contrast, ResNet never shows privacy risks before the 4th stage even if most additional layers from ResNet18 to ResNet152 belong to the 3rd stage. This phenomenon further explains the relationship between generalized features and privacy.

\paragraph{Does Prediction Disparity Exist Everywhere?} The answer is no! In short, privacy risks only exist \emph{in certain components without significant contributions to the model's generalizability}. Summarizing the observations above, we can draw the following insights of privacy and generalizability of an architecture:
\begin{itemize}
    \item \textbf{Privacy risk and generalizability are separable}: in standard training procedures, privacy-risky features are learned together with generalized features due to imperfect or inadequate designs in various aspects, such as loss function, label, and model architecture). Fortunately, they exist in different regions of neural networks, identifying the feasibility of decoupling them. 
    \item \textbf{Non- or less-generalized features lead to privacy risks}: Less-generalized features are usually learned in later layers. Although the increase in model computation capacity (i.e., increase in depth, channel size, and feature map size) brings the model effective improvements in generalizability, the surplus non- or less-generalized features put the model at higher privacy risks.
    \item \textbf{Privacy risk occurs in later stages}: The less-generalized features mainly exist in later stages, which is also consistent with the observation of \cite{baldock2021lendifficulty}. The more challenging to learn the feature is, the more possibly under-generalized the feature is. That is, privacy risk mostly exists in the later stage of the neural network.
\end{itemize}
These insights identify an issue in the current privacy defense approaches: they treat the model as a whole during training without distinguishing the privacy risks of various components. With no doubt, disturbing generalized privacy-safe features will lead to the unnecessary deterioration of the model's generalizability.
Hence, it is necessary to develop a more considerate training paradigm with regard to privacy.

\begin{algorithm}[t]
\caption{Privacy-Preserving Training Principle (PPTP)}
\label{alg:pptp}
\textbf{Input}: 
Training Dataset $\mathcal{D} = \{(x_i, y_i)\}_{i=1}^{N}$, Ordinarily Pre-Trained and Privacy-Risky Model $M_{pr}$ and its Parameters $\theta$, Privacy-Preserving Training Approach $f$, Training Epochs $E$, and Other Retraining Configurations $C$;
\\
\textbf{Output}: \makebox[.33\linewidth][s]{Privacy-Safe Model $M_{ps}$;}
\begin{algorithmic}[1] 
\STATE Split the model with parameters $\theta$ into Privacy-Safe and -Risk Layers, denoting their parameters as $\theta_{ps}$ and $\theta_{pr}$, respectively.
\STATE Freeze Privacy-Safe Parameters $\theta_{ps}$
\STATE Rewind Privacy-Risky Parameters $\theta_{pr}$
\
\FOR{$epoch$ \textbf{in} $\{1, 2, \cdots, E\}$}
\STATE Retrain the model with privacy-preserving training approach $f(\mathcal{D}, M_{pr}, C)$
\ENDFOR
\STATE Return privacy-safe model $M_{ps}$
\end{algorithmic}
\end{algorithm}

%% file: sec/approach.tex

\begin{figure*}[t]
     \centering
     \includegraphics[width=.85\linewidth]{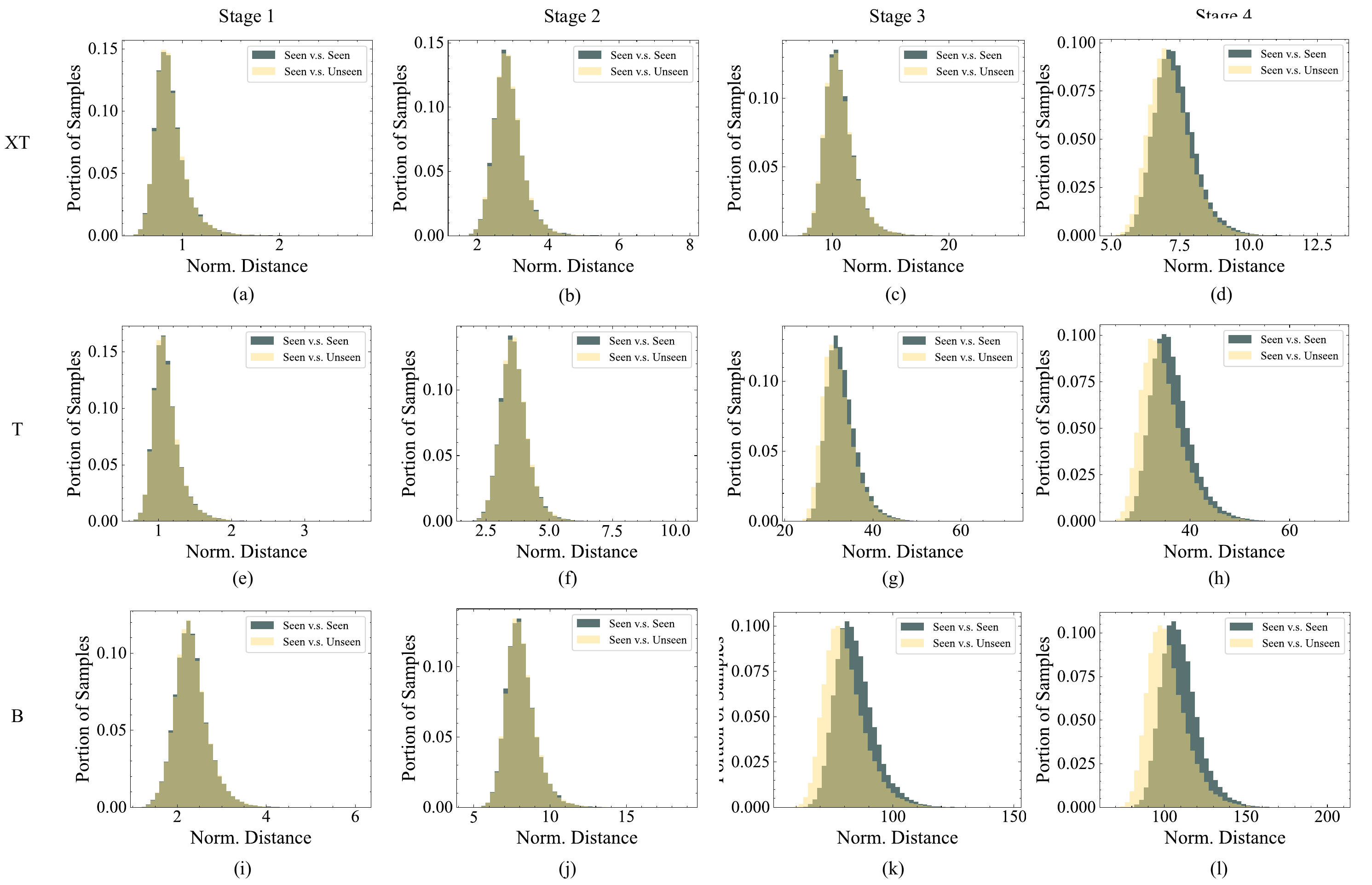}
\caption{Comparison of models in various depths. XT, T, and B denote ATM-XT, -T, and -B, respectively. (ATM, TinyImageNet)}
\label{fig:atm_depth}
\end{figure*}

\section{Privacy-Preserving Training Principle}
With the observations above in mind, we obtain the insight that a model does not need to update all weights to achieve better privacy-utility trade-offs. Instead, only privacy-risky layers can be retrained in a privacy-safe way. 
To achieve this, we propose a retraining approach, \emph{Privacy-Preserving Training Principle} (PPTP), to enable a model to minimize utility loss while obtaining privacy.
The approach is demonstrated in Alg.~\ref{alg:pptp}. In the algorithm, we first determine the privacy-risk and -safe layers. Then, \emph{the weights of privacy-safe layers are frozen since they have learned generalized and privacy-safe features.} After that, the weights of the privacy-risky layers are rewound for retraining. Because general training approaches do not take into account privacy criteria, these layers need to be trained with a privacy-defending approach to be privacy-safe. With our approach, PPTP, the model can maintain highly generalized features while reducing privacy-risky features. We empirically show its effectiveness by comparing differences between training with and without our approach in the next section.

%% file: sec/exp.tex
\section{Experiments}

\subsection{Experimental Settings}
 We evaluate our approach and others on CIFAR-100 \cite{cifar} and TinyImageNet \cite{tinyimagenet}. We evaluate our approaches on these two datasets to show the effectiveness on small and large datasets since extensive studies \cite{chen2018virtualsoftmax, yun2020cskd} have demonstrated that methods that work well on small datasets may have insignificant improvement on larger datasets. Besides, the layers to be frozen in a model may vary over different datasets, especially with different input dimensions or information domains. In other words, the model may produce privacy risks earlier or later depending on the dataset.
Data augmentation techniques, including random flipping \& cropping \cite{simonyan2015randomcrop}, are applied when training the model with a privacy-defending technique.
As for privacy attacks, we evaluate our approach and others on correctness-based MIAs \cite{yeom2020overfitting_robustness_malicious}, confidence-based MIAs \cite{yeom2018privacyoverfitting, song2019advexample, song2021systematic}, entropy-based MIAs \cite{shokri2017membership, song2021systematic}, modified-entropy-based MIAs \cite{song2021systematic}, and neural network based MIAs \cite{shokri2017membership, nasr2018advreg}. We include privacy training techniques such as adversarial regularization (\texttt{Adv-Reg}) \cite{nasr2018advreg}, SELENA \cite{tang2022selena}, and relaxed loss (\texttt{RelaxLoss}) \cite{chen2022relaxloss}. For each result, we execute three independent runs to ensure its stability. The hyper-parameters searching space of cited privacy defense approaches are shown in Table~\ref{tab:hyper-parameter_privacy}. Also, the main information of experimental software and hardware environment is presented in Table~\ref{tab:pc_conf}.

\begin{table}[t]
\centering
\caption{Hyper-parameters searching space.}\vskip 0.1in
  \resizebox{.85\linewidth}{!}{
  \begin{tabular}{c|ccc}
    \toprule
     Hyper-Parameter  & Adv-Reg      & SELENA     & RelaxLoss \\
    \midrule 
        $\alpha$      & $[1.0, 5.0]$ & N/A        & $[1.0, 3.0]$ \\
        $k$           & 3            & 25         & N/A \\
        $l$           & N/A          & $[3, 10]$  & N/A \\
    \bottomrule
  \end{tabular}
  }
\label{tab:hyper-parameter_privacy}
\end{table}
\begin{table}[t]
\centering
\caption{The information about computation environment.}\vskip 0.1in
  \resizebox{1.\linewidth}{!}{
  \begin{tabular}{c|cccccc}
    \toprule
                      & OS           & CPU      & RAM    & GPU           & CUDA & Pytroch \\
    \midrule 
        Information   & Ubuntu 22.04 & 96 cores & 460 GB & V100    & 12.1 & 2.1 \\
    \bottomrule
  \end{tabular}
  }
\label{tab:pc_conf}
\end{table}
\begin{figure}[t]
     \centering
     \includegraphics[width=1.\linewidth]{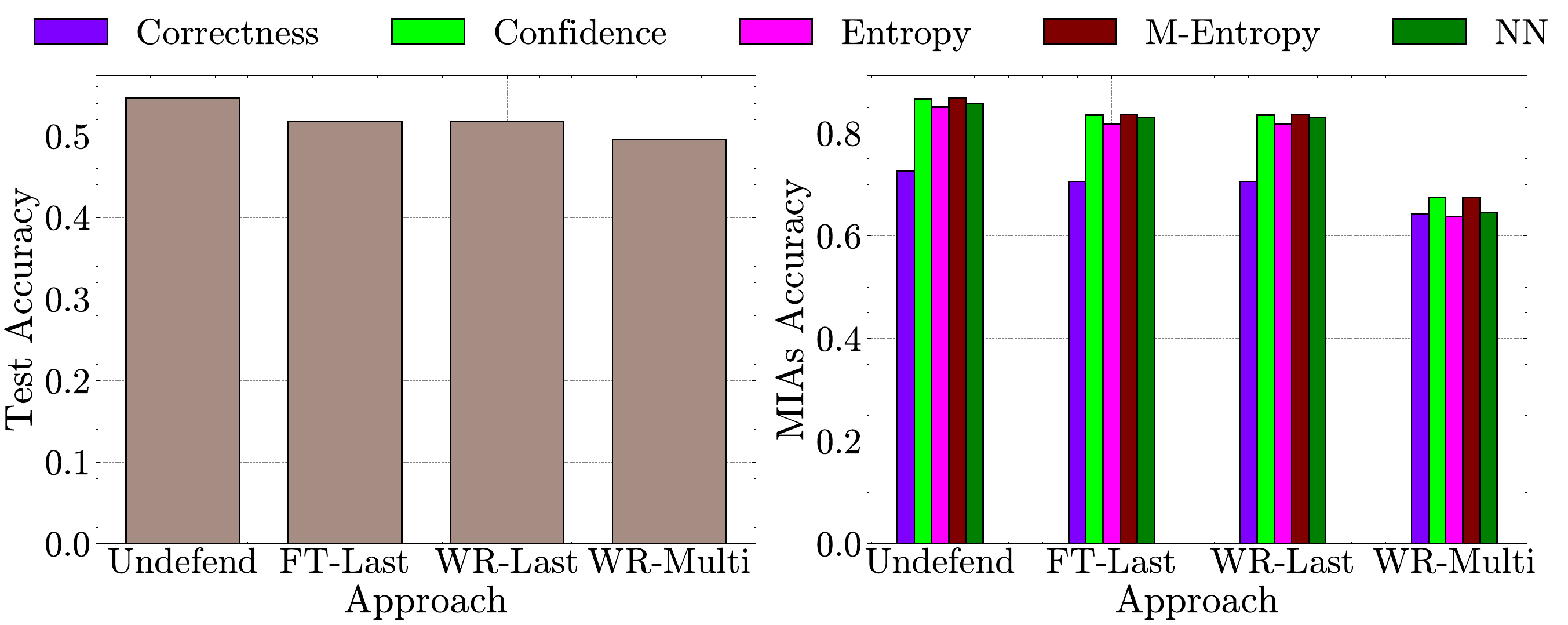}
\caption{Comparison of retraining the model with different designs (TinyImageNet, ResNet18).}
\label{fig:ablation_mia}
\end{figure}

\begin{figure*}[t!]
     \centering
     \includegraphics[width=1.0\linewidth]{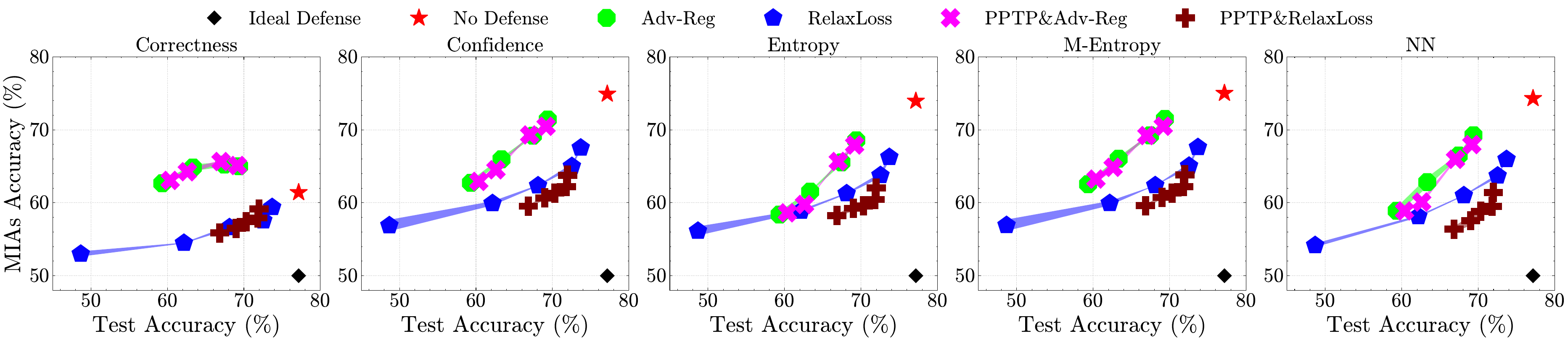}
\caption{Comparisons with existing privacy-preserving techniques (CIFAR100, ResNet18).}
\label{fig:mia_resnet18_c100}
\end{figure*}
\begin{figure*}[t!]
     \centering
     \includegraphics[width=1.0\linewidth]{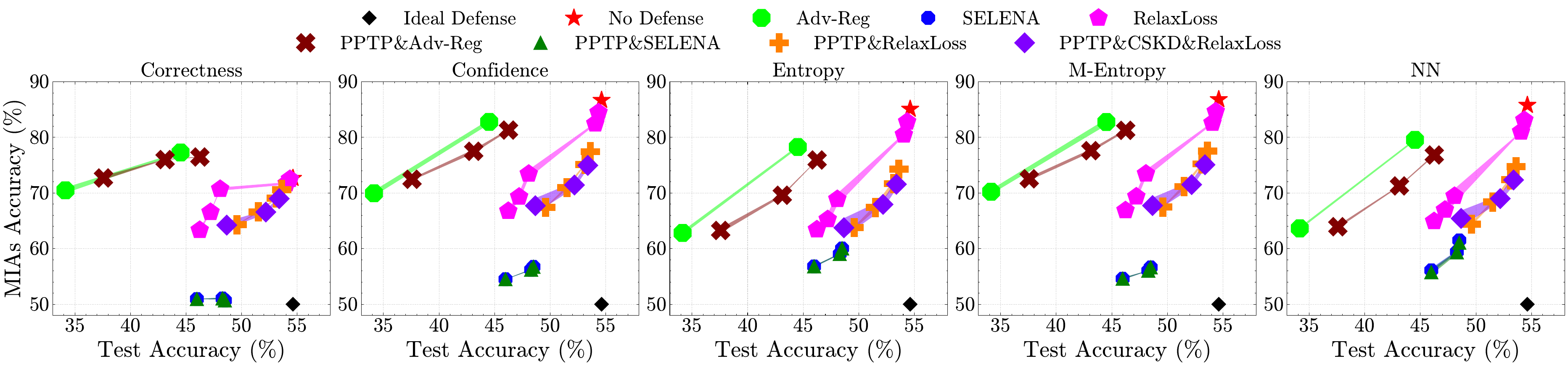}
\caption{Comparisons with existing privacy-preserving techniques (TinyImageNet, ResNet18).}
\label{fig:mia_resnet18_tmg}
\end{figure*}

\begin{figure*}[t!]
     \centering
     \includegraphics[width=.8\linewidth]{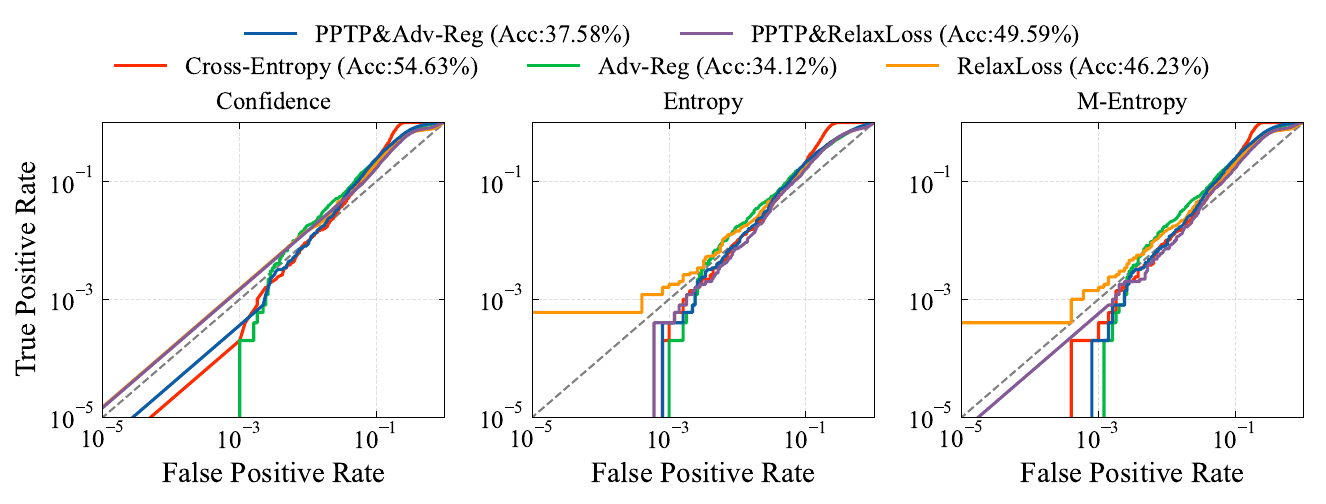}
\caption{AUC-ROC curve comparison of defense approaches with and without PPTP under various MIAs.}
\label{fig:mia_auc_tin_resnet18}
\end{figure*}

\begin{figure}
\centering
\includegraphics[width=1.\linewidth]{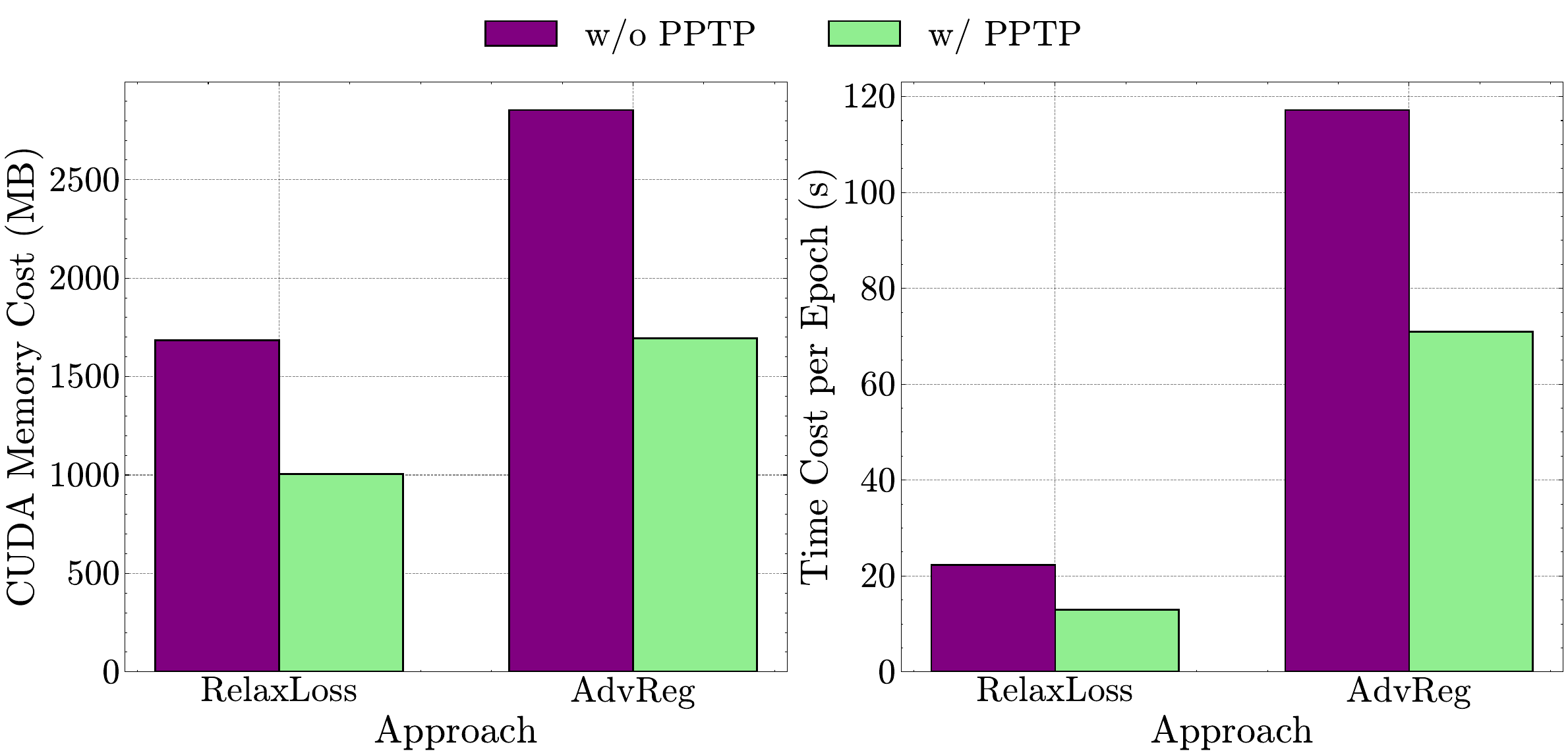}
\caption{Our proposed Privacy-Preserving Training Principle (PPTP) clearly decreases training costs in memory and time. (CIFAR100, ResNet18).}
\label{fig:train_cost}
\end{figure}

\begin{figure*}[t]
     \centering
     \includegraphics[width=.95\linewidth]{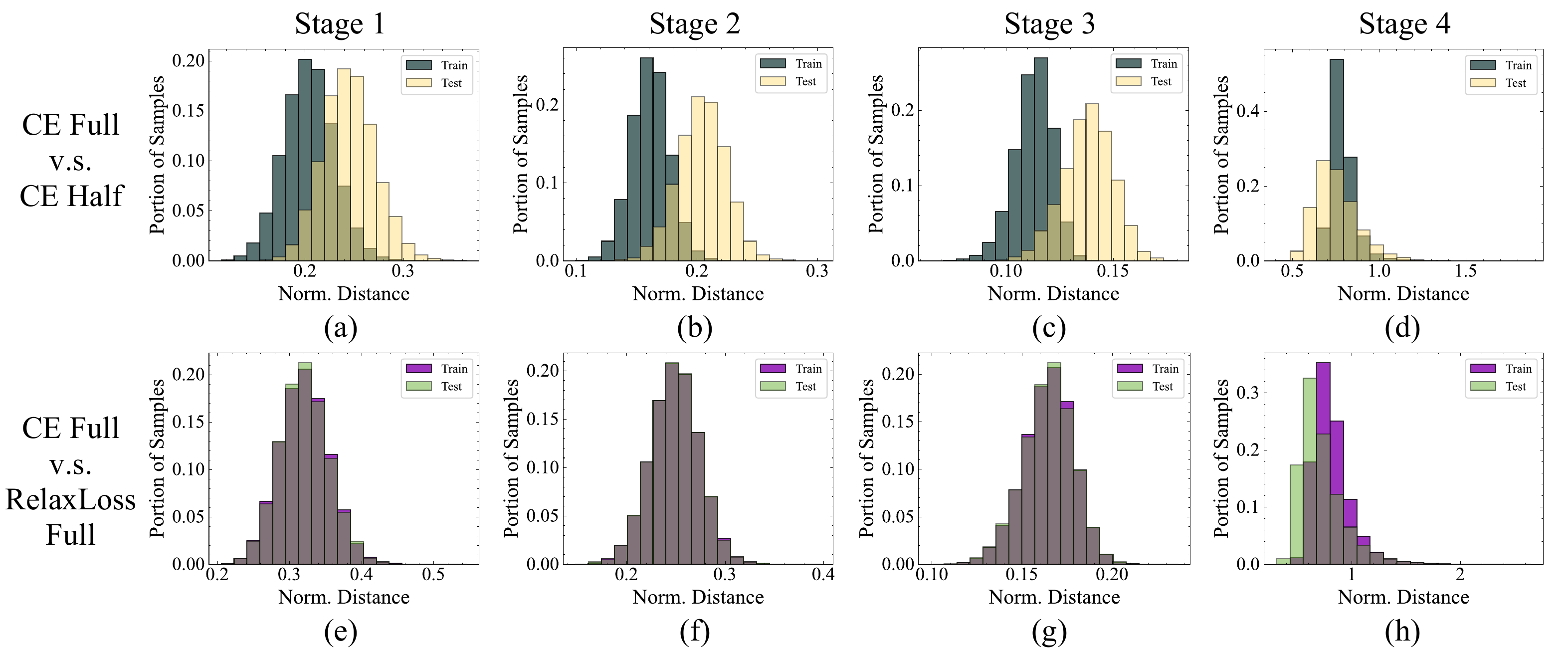}
\caption{The distribution differences between training full data with CE, half data with CE, and full data with RelaxLoss.}
\label{fig:ce_vs_rxl}
\end{figure*}

\subsection{Results and Discussions}
\subsubsection{Ablation Study}

To change the characteristics of a trained model, we can consider the two options, weight rewinding and fine-tuning. As seen in Algorithm~\ref{alg:pptp}, we choose rewinding and retraining to achieve more advanced utility-privacy trade-offs. Most privacy defense approaches, such as \cite{nasr2018advreg,shejwalkar2021dmp,chen2022relaxloss}, are designed for training from scratch. Thus rewinding is more adequate for them. 
Besides, it is necessary to check if retraining privacy-risky layers can help the model achieve better utility-privacy trade-offs. To examine it, we show the performance comparison between training from scratch, retraining privacy-risky layers, and retraining the classification layers using RelaxLoss. As seen in Figure~\ref{fig:ablation_mia}, retraining with the last layer only, regardless of whether through fine-tuning or weights rewinding, cannot effectively mitigate the model's privacy risks, while retraining multiple privacy-risky layers helps the model mitigate the privacy risks successfully.

\subsubsection{Comparisons with Other Approaches}
In CIFAR-100, we evaluate our approach with Adv-Reg and RelaxLoss. As shown in Figure~\ref{fig:mia_resnet18_c100}, both RelaxLoss and Adv-Reg show improvement when they are used to retrain the model with PPTP. However, the performance of correctness-based MIAs does not show significant improvement. PPTP retains the partially generalized features learned by CE to help the model maintain its utility, whereas it does not benefit the correctness alignment on train and test sets. Fortunately, however, this does not mean that PPTP cannot work with MIAs based on robustness or fairness. For instance, robustness issues of neural networks mainly exist in the final layer \cite{kirichenko2023robustlast}, meaning that the adversarial disparity can be also mitigated in the later layers to defend the robustness-based MIAs such as \cite{del2022leveraging}. 
We do not report the results of SELENA on this dataset since gradient explosions frequently occur on SELENA in the distillation stage (this phenomenon also occurs on TinyImageNet, although less frequently.)

In TinyImageNet, we evaluate our approach with Adv-Reg, SELENA, and RelaxLoss. Compared with results on CIFAR-100, the trends on TinyImageNet (Fig.~\ref{fig:mia_resnet18_tmg}) vary a lot. The Adv-Reg and RelaxLoss show more significant improvement once PPTP is applied. The results identify that our approach is effective in maintaining good performance on larger datasets.
An important factor is that we freeze much more weights in TinyImageNet than in CIFAR-100 because the disparity occurs in different layers (see Table~\ref{tab:stop_layer}). This also shows the potential of the model compression methods \cite{zhang2019your, zhang2022selfdistillation}, which can help the model achieve better utility performance at earlier stages, to be applied for privacy preservation. In contrast, the improvement in SELENA is very slight. We factor in that data augmentation is disabled in the distillation phase - if data augmentation was applied in this phase, the training time cost could have been over ten times greater than the current one. 
Additionally, it exhibits that our approach decouples the utility and privacy training. To demonstrate this characteristic, we apply class-wise self-knowledge distillation (\texttt{CSKD}) \cite{yun2020cskd} to pre-train the model. Compared with privacy training from scratch, pre-training with \texttt{CSKD} also shows improvements in utility at the same privacy level.  Finally, to further exhibit privacy improvement using our approach, we evaluate Adv-Reg and RelaxLoss with and without PPTP and plot the AUC-ROC curve \cite{carlini2022lira} for TinyImageNet. As shown in Figure~\ref{fig:mia_auc_tin_resnet18}, compared with original Adv-Reg and RelaxLoss, our approach achieves comparable or better privacy with higher testing accuracy, further identifying our approach's effectiveness.


\subsubsection{Training Cost}

The privacy training cost can be optimized via PPTP since PPTP freezes a large portion of the weights. We evaluate the GPU memory and time cost to show the efficiency benefit that PPTP enables through privacy training approaches. As shown in Figure~\ref{fig:train_cost}, the CUDA memory and time costs, evaluated on NVIDIA Tesla V100, clearly decrease after PPTP is applied. In particular, the actual training time cost is much less than the original training approaches (w/o PPTP) since retraining partial weights requires fewer epochs (less than half epochs) to converge.


\begin{table}[t]
\centering
\caption{Weight-freezing stopping layers}
\vskip 0.1in
\resizebox{.8\linewidth}{!}{
  \begin{tabular}{c|cc}
    \toprule
     Model & CIFAR-100 & TinyImageNet \\
    \midrule 
        ResNet18 & conv4\_x (stage 3) & conv5\_x (stage 4) \\
    \bottomrule
  \end{tabular}
  }
  \label{tab:stop_layer}
\end{table}

\begin{table}[t]
\centering
\caption{The performance comparison of different training data and training techniques. (ResNet18, TinyImageNet, data augmented)}
\vskip 0.1in
    \resizebox{1.0\linewidth}{!}{
  \begin{tabular}{c|cccc}
    \toprule
     Accuracy (\%) & CE \& Full Data & CE \& Half Data & RxL \& Full Data \\
    \midrule 
        Train & 99.98 & 99.99 & 79.68  \\
        Test  & 54.58 & 44.07 & 47.57  \\
    \bottomrule
  \end{tabular}
  }
\label{tab:cmp_data_loss}
\end{table}
%

\subsubsection{Feature Map Differences}
A significant factor that makes the model less generalized with a privacy-defending approach is that the feature representations in each model layer are quite different. As shown in Figure~\ref{fig:ce_vs_rxl}, the differences in feature maps between training with RelaxLoss and CE are more significant than training with different data (full vs. half). However, closer feature maps do not mean better generalizability (please refer to Table~\ref{tab:cmp_data_loss}). These identify that our approach can preserve the generalized features learned by CE at privacy-safe layers and help the privacy-risky layers become more privacy-safe to mitigate privacy leakage. 

%% file: sec/con.tex
\section{Conclusion}
In this paper, we observed that privacy vulnerability occurs at a portion of layers rather than the entire network. We underscore that the generalizability and privacy risks are decomposable since the well-generalized features and privacy-risky features exist in the different regions of the model. With this insight, we proposed Privacy-Preserving Training Principle (PPTP) to preserve generalizability along with privacy training. Through extensive empirical results, we showed that our approach enhances privacy with the proposed efficient training while not losing generalizability.

%% file: example_paper.bib
@inproceedings{devansh2017meminnn,
  title = 	 {A Closer Look at Memorization in Deep Networks},
  author =       {Devansh Arpit and Stanis{\l}aw Jastrz{\k{e}}bski and Nicolas Ballas and David Krueger and Emmanuel Bengio and Maxinder S. Kanwal and Tegan Maharaj and Asja Fischer and Aaron Courville and Yoshua Bengio and Simon Lacoste-Julien},
  booktitle = 	 {Proceedings of the 34th International Conference on Machine Learning},
  pages = 	 {233--242},
  year = 	 {2017},
  editor = 	 {Precup, Doina and Teh, Yee Whye},
  volume = 	 {70},
  series = 	 {Proceedings of Machine Learning Research},
  month = 	 {06--11 Aug},
  publisher =    {PMLR},
  url = 	 {https://proceedings.mlr.press/v70/arpit17a.html},
}

@inproceedings{chatterjee2018learnmemorization,
  title = 	 {Learning and Memorization},
  author =       {Chatterjee, Satrajit},
  booktitle = 	 {Proceedings of the 35th International Conference on Machine Learning},
  pages = 	 {755--763},
  year = 	 {2018},
  editor = 	 {Dy, Jennifer and Krause, Andreas},
  volume = 	 {80},
  series = 	 {Proceedings of Machine Learning Research},
  month = 	 {10--15 Jul},
  publisher =    {PMLR},
  url = 	 {https://proceedings.mlr.press/v80/chatterjee18a.html},
}

@inproceedings{baldock2021lendifficulty,
 author = {Baldock, Robert and Maennel, Hartmut and Neyshabur, Behnam},
 booktitle = {Advances in Neural Information Processing Systems},
 editor = {M. Ranzato and A. Beygelzimer and Y. Dauphin and P.S. Liang and J. Wortman Vaughan},
 pages = {10876--10889},
 publisher = {Curran Associates, Inc.},
 title = {Deep Learning Through the Lens of Example Difficulty},
 url = {https://proceedings.neurips.cc/paper_files/paper/2021/file/5a4b25aaed25c2ee1b74de72dc03c14e-Paper.pdf},
 volume = {34},
 year = {2021}
}

@inproceedings{stephenson2021geometry,
title={On the geometry of generalization and memorization in deep neural networks},
author={Cory Stephenson and suchismita padhy and Abhinav Ganesh and Yue Hui and Hanlin Tang and SueYeon Chung},
booktitle={International Conference on Learning Representations},
year={2021},
url={https://openreview.net/forum?id=V8jrrnwGbuc}
}

@inproceedings{carlini2022privacyonion,
title={The Privacy Onion Effect: Memorization is Relative},
author={Nicholas Carlini and Matthew Jagielski and Chiyuan Zhang and Nicolas Papernot and Andreas Terzis and Florian Tramer},
booktitle={Advances in Neural Information Processing Systems},
editor={Alice H. Oh and Alekh Agarwal and Danielle Belgrave and Kyunghyun Cho},
year={2022},
url={https://openreview.net/forum?id=ErUlLrGaVEU}
}

@inproceedings{carlini2022lira,
  title={Membership inference attacks from first principles},
  author={Carlini, Nicholas and Chien, Steve and Nasr, Milad and Song, Shuang and Terzis, Andreas and Tramer, Florian},
  booktitle={2022 IEEE Symposium on Security and Privacy (SP)},
  pages={1897--1914},
  year={2022},
  organization={IEEE}
}

@inproceedings{fredrikson2015mi,
  title={Model inversion attacks that exploit confidence information and basic countermeasures},
  author={Fredrikson, Matt and Jha, Somesh and Ristenpart, Thomas},
  booktitle={Proceedings of the 22nd ACM SIGSAC conference on computer and communications security},
  pages={1322--1333},
  year={2015}
}

@inproceedings{tramer2016me,
  title={Stealing machine learning models via prediction $\{$APIs$\}$},
  author={Tram{\`e}r, Florian and Zhang, Fan and Juels, Ari and Reiter, Michael K and Ristenpart, Thomas},
  booktitle={25th USENIX security symposium (USENIX Security 16)},
  pages={601--618},
  year={2016}
}

@inproceedings{shokri2017membership,
  title={Membership inference attacks against machine learning models},
  author={Shokri, Reza and Stronati, Marco and Song, Congzheng and Shmatikov, Vitaly},
  booktitle={2017 IEEE symposium on security and privacy (SP)},
  pages={3--18},
  year={2017},
  organization={IEEE}
}

@inproceedings{chen2020ganleaks,
author = {Dingfan Chen and Ning Yu and Yang Zhang and Mario Fritz},
title = {GAN-Leaks: A Taxonomy of Membership Inference Attacks against Generative Models},
booktitle = {ACM Conference on Computer and Communications Security (CCS)},
year = {2020}
}

@misc{zou2020miatransfer,
      title={Privacy Analysis of Deep Learning in the Wild: Membership Inference Attacks against Transfer Learning}, 
      author={Yang Zou and Zhikun Zhang and Michael Backes and Yang Zhang},
      year={2020},
      eprint={2009.04872},
      archivePrefix={arXiv},
      primaryClass={cs.CR},
      url={https://arxiv.org/abs/2009.04872}, 
}

@article{wu2024rethinkmiatransfer,
  author={Wu, Cong and Chen, Jing and Fang, Qianru and He, Kun and Zhao, Ziming and Ren, Hao and Xu, Guowen and Liu, Yang and Xiang, Yang},
  journal={IEEE Transactions on Information Forensics and Security}, 
  title={Rethinking Membership Inference Attacks Against Transfer Learning}, 
  year={2024},
  volume={19},
  number={},
  pages={6441-6454},
  doi={10.1109/TIFS.2024.3413592}}

@inproceedings{song2019advexample,
author = {Song, Liwei and Shokri, Reza and Mittal, Prateek},
title = {Privacy Risks of Securing Machine Learning Models against Adversarial Examples},
year = {2019},
isbn = {9781450367479},
publisher = {Association for Computing Machinery},
address = {New York, NY, USA},
url = {https://doi.org/10.1145/3319535.3354211},
doi = {10.1145/3319535.3354211},
booktitle = {Proceedings of the 2019 ACM SIGSAC Conference on Computer and Communications Security},
pages = {241–257},
numpages = {17},
location = {London, United Kingdom},
series = {CCS '19}
}

@inproceedings{yeom2018privacyoverfitting,
  title={Privacy risk in machine learning: Analyzing the connection to overfitting},
  author={Yeom, Samuel and Giacomelli, Irene and Fredrikson, Matt and Jha, Somesh},
  booktitle={2018 IEEE 31st computer security foundations symposium (CSF)},
  pages={268--282},
  year={2018},
  organization={IEEE}
}

@inproceedings{choquette2021labelonlymia,
  title={Label-only membership inference attacks},
  author={Choquette-Choo, Christopher A and Tramer, Florian and Carlini, Nicholas and Papernot, Nicolas},
  booktitle={International conference on machine learning},
  pages={1964--1974},
  year={2021},
}

@article{yeom2020overfitting_robustness_malicious,
  title={Overfitting, robustness, and malicious algorithms: A study of potential causes of privacy risk in machine learning},
  author={Yeom, Samuel and Giacomelli, Irene and Menaged, Alan and Fredrikson, Matt and Jha, Somesh},
  journal={Journal of Computer Security},
  volume={28},
  number={1},
  pages={35--70},
  year={2020},
  publisher={IOS Press}
}

@inproceedings{del2022leveraging,
  title={Leveraging adversarial examples to quantify membership information leakage},
  author={Del Grosso, Ganesh and Jalalzai, Hamid and Pichler, Georg and Palamidessi, Catuscia and Piantanida, Pablo},
  booktitle={Proceedings of the IEEE/CVF Conference on Computer Vision and Pattern Recognition},
  pages={10399--10409},
  year={2022}
}

@inproceedings{song2021systematic,
  title={Systematic evaluation of privacy risks of machine learning models},
  author={Song, Liwei and Mittal, Prateek},
  booktitle={30th USENIX Security Symposium (USENIX Security 21)},
  pages={2615--2632},
  year={2021}
}

@inproceedings{abadi2016dpsgd,
  title={Deep learning with differential privacy},
  author={Abadi, Martin and Chu, Andy and Goodfellow, Ian and McMahan, H Brendan and Mironov, Ilya and Talwar, Kunal and Zhang, Li},
  booktitle={Proceedings of the 2016 ACM SIGSAC conference on computer and communications security},
  pages={308--318},
  year={2016}
}

@inproceedings{jia2019memguard,
  title={Memguard: Defending against black-box membership inference attacks via adversarial examples},
  author={Jia, Jinyuan and Salem, Ahmed and Backes, Michael and Zhang, Yang and Gong, Neil Zhenqiang},
  booktitle={Proceedings of the 2019 ACM SIGSAC conference on computer and communications security},
  pages={259--274},
  year={2019}
}

@inproceedings{yang2023purifier,
  title={Purifier: Defending Data Inference Attacks via Transforming Confidence Scores},
  author={Yang, Ziqi and Wang, Lijin and Yang, Da and Wan, Jie and Zhao, Ziming and Chang, Ee-Chien and Zhang, Fan and Ren, Kui},
  booktitle={Proceedings of the AAAI Conference on Artificial Intelligence},
  volume={37},
  number={9},
  pages={10871--10879},
  year={2023}
}

@inproceedings{nasr2018advreg,
  title={Machine learning with membership privacy using adversarial regularization},
  author={Nasr, Milad and Shokri, Reza and Houmansadr, Amir},
  booktitle={Proceedings of the 2018 ACM SIGSAC conference on computer and communications security},
  pages={634--646},
  year={2018}
}

@inproceedings{shejwalkar2021dmp,
  title={Membership privacy for machine learning models through knowledge transfer},
  author={Shejwalkar, Virat and Houmansadr, Amir},
  booktitle={Proceedings of the AAAI conference on artificial intelligence},
  volume={35},
  number={11},
  pages={9549--9557},
  year={2021}
}

@inproceedings{wang2021pruning,
  title     = {Against Membership Inference Attack: Pruning is All You Need},
  author    = {Wang, Yijue and Wang, Chenghong and Wang, Zigeng and Zhou, Shanglin and Liu, Hang and Bi, Jinbo and Ding, Caiwen and Rajasekaran, Sanguthevar},
  booktitle = {Proceedings of the Thirtieth International Joint Conference on
               Artificial Intelligence, {IJCAI-21}},
  pages     = {3141--3147},
  year      = {2021}
}

@inproceedings{li2021mixupmmd,
  title={Membership inference attacks and defenses in classification models},
  author={Li, Jiacheng and Li, Ninghui and Ribeiro, Bruno},
  booktitle={Proceedings of the Eleventh ACM Conference on Data and Application Security and Privacy},
  pages={5--16},
  year={2021}
}

@inproceedings{yuan2022samia,
  title={Membership inference attacks and defenses in neural network pruning},
  author={Yuan, Xiaoyong and Zhang, Lan},
  booktitle={31st USENIX Security Symposium (USENIX Security 22)},
  pages={4561--4578},
  year={2022}
}

@inproceedings{tang2022selena,
  title={Mitigating Membership Inference Attacks by Self-Distillation Through a Novel Ensemble Architecture},
  author={Tang, Xinyu and Mahloujifar, Saeed and Song, Liwei and Shejwalkar, Virat and Nasr, Milad and Houmansadr, Amir and Mittal, Prateek},
  booktitle = {31st {USENIX} Security Symposium ({USENIX} Security)},
  year={2022}
}

@inproceedings{chen2022relaxloss,
  title={RelaxLoss: Defending Membership Inference Attacks without Losing Utility},
  author={Dingfan Chen and Ning Yu and Mario Fritz},
  booktitle={International Conference on Learning Representations},
  year={2022}
}

@inproceedings{tan2023blessing,
  title = 	 {A Blessing of Dimensionality in Membership Inference through Regularization},
  author =       {Tan, Jasper and LeJeune, Daniel and Mason, Blake and Javadi, Hamid and Baraniuk, Richard G.},
  booktitle = 	 {Proceedings of The 26th International Conference on Artificial Intelligence and Statistics},
  pages = 	 {10968--10993},
  year = 	 {2023},
  editor = 	 {Ruiz, Francisco and Dy, Jennifer and van de Meent, Jan-Willem},
  volume = 	 {206},
  series = 	 {Proceedings of Machine Learning Research},
  month = 	 {25--27 Apr},
  publisher =    {PMLR},
}

@inproceedings{chen2023hamp,
  title={Overconfidence is a Dangerous Thing: Mitigating Membership Inference Attacks by Enforcing Less Confident Prediction}, 
  author={Chen, Zitao and Pattabiraman, Karthik},
  booktitle = {Network and Distributed System Security (NDSS) Symposium},
  year={2024}
}

@inproceedings{fang2024crl,
title={Center-Based Relaxed Learning Against Membership Inference Attacks},
author={Xingli Fang and Jung-Eun Kim},
booktitle={The 40th Conference on Uncertainty in Artificial Intelligence},
year={2024},
url={https://openreview.net/forum?id=unlWrunFjg}
}

@misc{fang2024srcm,
      title={Representation Magnitude has a Liability to Privacy Vulnerability}, 
      author={Xingli Fang and Jung-Eun Kim},
      year={2024},
      eprint={2407.16164},
      archivePrefix={arXiv},
      primaryClass={cs.LG},
      url={https://arxiv.org/abs/2407.16164}, 
}

@inproceedings{liu2024ccl,
  title = 	 {Mitigating Privacy Risk in Membership Inference by Convex-Concave Loss},
  author =       {Liu, Zhenlong and Feng, Lei and Zhuang, Huiping and Cao, Xiaofeng and Wei, Hongxin},
  booktitle = 	 {Proceedings of the 41st International Conference on Machine Learning},
  pages = 	 {30998--31014},
  year = 	 {2024},
  editor = 	 {Salakhutdinov, Ruslan and Kolter, Zico and Heller, Katherine and Weller, Adrian and Oliver, Nuria and Scarlett, Jonathan and Berkenkamp, Felix},
  volume = 	 {235},
  series = 	 {Proceedings of Machine Learning Research},
  month = 	 {21--27 Jul},
  publisher =    {PMLR},
  url = 	 {https://proceedings.mlr.press/v235/liu24q.html},
}

@inproceedings{li2024mist,
  title={$\{$MIST$\}$: Defending against membership inference attacks through $\{$Membership-Invariant$\}$ subspace training},
  author={Li, Jiacheng and Li, Ninghui and Ribeiro, Bruno},
  booktitle={33rd USENIX Security Symposium (USENIX Security 24)},
  pages={2387--2404},
  year={2024}
}

@inproceedings{zhao2025does_syn_data_protect_privacy,
title={Does Training with Synthetic Data Truly Protect Privacy?},
author={Yunpeng Zhao and Jie Zhang},
booktitle={The Thirteenth International Conference on Learning Representations},
year={2025},
url={https://openreview.net/forum?id=C8niXBHjfO}
}

@inproceedings{shang2025defending_mias_iteratively_prune_dnn,
  title={Defending Against Membership Inference Attacks on Iteratively Pruned Deep Neural Networks.},
  author={Shang, Jing and Wang, Jian and Wang, Kailun and Liu, Jiqiang and Jiang, Nan and Armanuzzaman, Md and Zhao, Ziming},
  booktitle={NDSS},
  year={2025}
}

@inproceedings{zhang2024archprivacy,
  title={How Does a Deep Learning Model Architecture Impact Its Privacy? A Comprehensive Study of Privacy Attacks on $\{$CNNs$\}$ and Transformers},
  author={Zhang, Guangsheng and Liu, Bo and Tian, Huan and Zhu, Tianqing and Ding, Ming and Zhou, Wanlei},
  booktitle={33rd USENIX Security Symposium (USENIX Security 24)},
  pages={6795--6812},
  year={2024}
}

@inproceedings{yu2021howdoesdataaug,
  title={How does data augmentation affect privacy in machine learning?},
  author={Yu, Da and Zhang, Huishuai and Chen, Wei and Yin, Jian and Liu, Tie-Yan},
  booktitle={Proceedings of the AAAI Conference on Artificial Intelligence},
  volume={35},
  number={12},
  pages={10746--10753},
  year={2021}
}

@inproceedings{kaya2021whendataaug,
  title = 	 {When Does Data Augmentation Help With Membership Inference Attacks?},
  author =       {Kaya, Yigitcan and Dumitras, Tudor},
  booktitle = 	 {Proceedings of the 38th International Conference on Machine Learning},
  pages = 	 {5345--5355},
  year = 	 {2021},
  editor = 	 {Meila, Marina and Zhang, Tong},
  volume = 	 {139},
  series = 	 {Proceedings of Machine Learning Research},
  month = 	 {18--24 Jul},
  publisher =    {PMLR},
  url = 	 {https://proceedings.mlr.press/v139/kaya21a.html}
}

@inproceedings{bourtoule2021mul,
  title={Machine unlearning},
  author={Bourtoule, Lucas and Chandrasekaran, Varun and Choquette-Choo, Christopher A and Jia, Hengrui and Travers, Adelin and Zhang, Baiwu and Lie, David and Papernot, Nicolas},
  booktitle={2021 IEEE Symposium on Security and Privacy (SP)},
  pages={141--159},
  year={2021},
  organization={IEEE}
}

@misc{goodfellow2015advattack,
      title={Explaining and Harnessing Adversarial Examples}, 
      author={Ian J. Goodfellow and Jonathon Shlens and Christian Szegedy},
      year={2015},
      eprint={1412.6572},
      archivePrefix={arXiv},
      primaryClass={stat.ML},
      url={https://arxiv.org/abs/1412.6572}, 
}

@article{mehrabi2021fairnesssurvey,
author = {Mehrabi, Ninareh and Morstatter, Fred and Saxena, Nripsuta and Lerman, Kristina and Galstyan, Aram},
title = {A Survey on Bias and Fairness in Machine Learning},
year = {2021},
issue_date = {July 2022},
publisher = {Association for Computing Machinery},
address = {New York, NY, USA},
volume = {54},
number = {6},
issn = {0360-0300},
url = {https://doi.org/10.1145/3457607},
doi = {10.1145/3457607},
journal = {ACM Comput. Surv.},
month = {jul},
articleno = {115},
numpages = {35}
}

@inproceedings{wei2023atm,
  title={Active token mixer},
  author={Wei, Guoqiang and Zhang, Zhizheng and Lan, Cuiling and Lu, Yan and Chen, Zhibo},
  booktitle={Proceedings of the AAAI Conference on Artificial Intelligence},
  volume={37},
  number={3},
  pages={2759--2767},
  year={2023}
}

@inproceedings{simonyan2015vgg,
  author       = {Karen Simonyan and Andrew Zisserman},
  title        = {Very Deep Convolutional Networks for Large-Scale Image Recognition},
  booktitle    = {3rd International Conference on Learning Representations, {ICLR} 2015,
                  San Diego, CA, USA, May 7-9, 2015, Conference Track Proceedings},
  year         = {2015}
}

@inproceedings{he2016resnet,
  title={Deep residual learning for image recognition},
  author={He, Kaiming and Zhang, Xiangyu and Ren, Shaoqing and Sun, Jian},
  booktitle={Proceedings of the IEEE conference on computer vision and pattern recognition},
  pages={770--778},
  year={2016}
}

@article{cortes1995svm,
  title={Support-vector networks},
  author={Cortes, Corinna and Vapnik, Vladimir},
  journal={Machine learning},
  volume={20},
  pages={273--297},
  year={1995},
  publisher={Springer}
}

@inproceedings{chen2018virtualsoftmax,
 author = {Chen, Binghui and Deng, Weihong and Shen, Haifeng},
 booktitle = {Advances in Neural Information Processing Systems},
 editor = {S. Bengio and H. Wallach and H. Larochelle and K. Grauman and N. Cesa-Bianchi and R. Garnett},
 pages = {},
 publisher = {Curran Associates, Inc.},
 title = {Virtual Class Enhanced Discriminative Embedding Learning},
 url = {https://proceedings.neurips.cc/paper_files/paper/2018/file/d79aac075930c83c2f1e369a511148fe-Paper.pdf},
 volume = {31},
 year = {2018}
}

@inproceedings{yun2020cskd,
  title={Regularizing class-wise predictions via self-knowledge distillation},
  author={Yun, Sukmin and Park, Jongjin and Lee, Kimin and Shin, Jinwoo},
  booktitle={Proceedings of the IEEE/CVF conference on computer vision and pattern recognition},
  pages={13876--13885},
  year={2020}
}

@inproceedings{kirichenko2023robustlast,
title={Last Layer Re-Training is Sufficient for Robustness to Spurious Correlations},
author={Polina Kirichenko and Pavel Izmailov and Andrew Gordon Wilson},
booktitle={The Eleventh International Conference on Learning Representations },
year={2023},
url={https://openreview.net/forum?id=Zb6c8A-Fghk}
}

@inproceedings{shafahi2019advfree,
 author = {Shafahi, Ali and Najibi, Mahyar and Ghiasi, Mohammad Amin and Xu, Zheng and Dickerson, John and Studer, Christoph and Davis, Larry S and Taylor, Gavin and Goldstein, Tom},
 booktitle = {Advances in Neural Information Processing Systems},
 editor = {H. Wallach and H. Larochelle and A. Beygelzimer and F. d\textquotesingle Alch\'{e}-Buc and E. Fox and R. Garnett},
 pages = {},
 publisher = {Curran Associates, Inc.},
 title = {Adversarial training for free!},
 url = {https://proceedings.neurips.cc/paper_files/paper/2019/file/7503cfacd12053d309b6bed5c89de212-Paper.pdf},
 volume = {32},
 year = {2019}
}

@misc{simonyan2015randomcrop,
      title={Very Deep Convolutional Networks for Large-Scale Image Recognition}, 
      author={Karen Simonyan and Andrew Zisserman},
      year={2015},
      eprint={1409.1556},
      archivePrefix={arXiv},
      primaryClass={cs.CV}
}

@misc{cifar,
  title={Learning multiple layers of features from tiny images},
  author={Krizhevsky, Alex and Hinton, Geoffrey and others},
  year={2009},
  publisher={Toronto, ON, Canada}
}

@misc{tinyimagenet,
  title={Tiny imagenet visual recognition challenge},
  author={Le, Ya and Yang, Xuan},
  journal={CS 231N},
  volume={7},
  number={7},
  pages={3},
  year={2015}
}

@inproceedings{zhang2019your,
  title={Be your own teacher: Improve the performance of convolutional neural networks via self distillation},
  author={Zhang, Linfeng and Song, Jiebo and Gao, Anni and Chen, Jingwei and Bao, Chenglong and Ma, Kaisheng},
  booktitle={Proceedings of the IEEE/CVF international conference on computer vision},
  pages={3713--3722},
  year={2019}
}

@article{zhang2022selfdistillation,
  author={Zhang, Linfeng and Bao, Chenglong and Ma, Kaisheng},
  journal={IEEE Transactions on Pattern Analysis and Machine Intelligence}, 
  title={Self-Distillation: Towards Efficient and Compact Neural Networks}, 
  year={2022},
  volume={44},
  number={8},
  pages={4388-4403}
}


%% file: main.bib
@inproceedings{fang2026learnability,
title={Learnability and Privacy Vulnerability are Entangled in a Few Critical Weights},
author={Fang, Xingli and Kim, Jung-Eun},
booktitle={The Fourteenth International Conference on Learning Representations (ICLR)},
year={2026},
url={https://openreview.net/forum?id=J2gI585XDK}
}

@misc{fang2025trustworthyaisafetybias,
      title={Trustworthy AI: Safety, Bias, and Privacy -- A Survey}, 
      author={Xingli Fang and Jianwei Li and Varun Mulchandani and Jung-Eun Kim},
      year={2025},
      eprint={2502.10450},
      archivePrefix={arXiv},
      primaryClass={cs.CR},
      url={https://arxiv.org/abs/2502.10450}, 
}

@inproceedings{Cuong2022NeurIPS,
 author = {Cuong Tran and Ferdinando Fioretto and Jung-Eun Kim and Rakshit Naidu},
 booktitle = {Advances in Neural Information Processing Systems},
 title = {Pruning has a disparate impact on model accuracy},
 url = {https://openreview.net/forum?id=11nMVZK0WYM},
 year = {2022}
}

@inproceedings{Bellam2025QFairness,
 author = {Abhimanyu Bellam and Jung-Eun Kim},
 booktitle = {Advances in Neural Information Processing Systems},
 title = {Explaining How Quantization Disparately Skews a Model},
 url = {https://arxiv.org/abs/2509.07222},
 year = {2025}
}

@inproceedings{Bellam2025QFITMLNeurIPS,
 author = {Abhimanyu Bellam and Jung-Eun Kim},
 booktitle = {FITML at NeurIPS},
 title = {Explaining How Quantization Disparately Skews a Model},
 url = {https://openreview.net/pdf?id=C0A0otZL2a},
 year = {2024}
}

@inproceedings{Mulchandani2025ICLR,
title={Severing Spurious Correlations with Data Pruning},
author={Varun Mulchandani and Kim, Jung-Eun},
booktitle={The Fourteenth International Conference on Learning Representations (ICLR)},
year={2025},
url={https://openreview.net/forum?id=Bk13Qfu8Ru}
}
